\RequirePackage{fix-cm}
\documentclass[twocolumn]{svjour3}          
\smartqed  
\usepackage{graphicx}
\usepackage{marvosym}
\usepackage{amstext,amssymb,amsbsy}
\usepackage{subfigure}
\usepackage{pstricks}
\usepackage{pst-node} 
\newcommand{\xor}{Exclusive-OR}
\newcommand{\ior}{Inclusive-OR}
\newcommand{\gaOne}{\emph{genetic\_search\_one}}
\newcommand{\gaOneSans}{\emph{mutation\_search\_one}}
\newcommand{\bsOne}{\emph{blind\_search\_one}}
\newcommand{\plotWidth}{.4\textwidth} 
\psset{arrowsize=5pt}   

\newcommand{\oneVector}[1]{%
\left[
\begin{array}{c}
#1
\end{array} \right]
}

\newcommand{\twoVector}[2]{%
\left[
\begin{array}{c}
#1 \\
#2
\end{array} \right]
}

\newcommand{\fourVector}[4]{%
\left[
\begin{array}{c}
#1 \\
#2 \\
#3 \\
#4 \\
\end{array} \right]
}
\newcommand{\toprule}{\hline\noalign{\smallskip}}
\newcommand{\midrule}{\noalign{\smallskip}\hline\noalign{\smallskip}}
\newcommand{\bottomrule}{\noalign{\smallskip}\hline}
\journalname{Evolutionary Intelligence}
\begin{document}

\title{Evolving A-Type Artificial Neural Networks}


\author{Ewan Orr \and
        Ben Martin 
}


\institute{Ben Martin \at
              Department of Mathematics and Statistics \\
              University of Canterbury\\
              Private Bag 4800\\
              Christchurch 8140\\
              New Zealand\\
              Tel.: +64 3 364 2987 ext 7687 \\
              Fax: +64 3 364 2587 \\
              \email{b.martin@math.canterbury.ac.nz}           
           \and
           Ewan Orr \at
              Department of Physics and Astronomy \\
              University of Canterbury\\
              New Zealand\\
              \email{ewan.orr@canterbury.ac.nz}           
}

\date{Received: date / Accepted: date}

\maketitle

\begin{abstract}
We investigate Turing's notion of an A-type artificial neural network.  We study a
refinement of Turing's original idea, motivated by work of Teuscher, Bull, Preen and Copeland.
Our A-types can process binary data by accepting and outputting sequences of binary
vectors; hence we can associate a function to an A-type, and we say the A-type
{\em represents} the function.  There are two modes of data processing: clamped and sequential.
We describe an evolutionary algorithm, involving graph-theoretic manipulations
of A-types, which searches for A-types representing a given function.  The algorithm uses both
 mutation and crossover operators. We implemented
the algorithm and applied it to three benchmark tasks.  We found that the algorithm performed much
better than a random search.  For two out of the three tasks, the algorithm with crossover
performed better than a mutation-only version.
\end{abstract}

\section{Introduction}
\label{intro}

In this paper we
report on our investigations into one of Alan Turing's contributions to artificial intelligence.  In 1948
Turing introduced a type of artificial neural network (ANN), which he called an A-type unorganised
machine.  Motivated by his work and by work of Teuscher, Bull, Preen and Copeland (see
Section \ref{sec:history}), we study a refinement of Turing's notion, which we call an A-type.

A-types can be used to process binary data: with suitable conventions involving input and output
 nodes, one can input a string of binary vectors into an A-type and receive a string of binary
 vectors as output.  Hence we can associate a function to an A-type; we say that the A-type
 {\em represents} this function.  We devised an evolutionary algorithm (EA) to design an
 A-type that represents a given function $f$.  We use a graph-based representation for A-types,
 and our EA---and in particular, our crossover operator---is based on graph-theoretic ideas.
We implemented our algorithm and applied it to three benchmark problems.

Turing's research on A-types is of great historical interest.  As the centenary of his birth approaches,
it is fitting to apply modern ideas---such as the theory of non-linear dynamical systems---to
his ground-breaking work.  A-types are an excellent test-bed for these ideas: they are composed of
neurons with a very simple firing rule and they are easy to program, but they are also powerful.
In this paper we adapt some existing ideas such as graph-based chromosomes and sequential input to
the setting of A-types.  The use of sequential input mode here brings up some new problems which
motivated us to introduce a new kind of neuron, delay nodes, not originally envisaged by Turing (see
Section~\ref{sec:delaysNeeded}).  Our graph-based EA works in the settings of both sequential and
clamped input.

In Section~\ref{sec:history} we give a brief survey of previous work on A-types.  In Section~\ref{sec:ours}
we present our interpretation of A-types, and we describe our EA in Section~\ref{sec:evolution}.
Section~\ref{sec:simulations} contains the results of our experimental work.

Our investigations are mainly at the proof-of-concept level.  Our EA has many parameters and we chose their values in an ad-hoc fashion to ensure that solutions were quickly discovered reasonably often; we did not search systematically for the optimum values (but see Sections~\ref{sec:par_bias}~and~\ref{sec:evolving_ops}).

\section{Historical Background and Previous Work}\label{sec:history}


In 1948 Turing wrote the pioneering technical report
 \emph{Intelligent Machinery}~\cite{turing1948}. In this report
 he introduced a type of ANN which he called an A-type unorganised
 machine\footnote{This was seemingly independent~\cite[p408]{copelandOne} of
 the 1943 paper~\cite{mcculloch1943logical} of McCulloch and Pitts in which ANNs were
 first introduced.}. This ANN is discrete, synchronously updated and, in
 general, recurrent. It is composed of basic and identical neurons (or nodes) each of
which performs the Boolean operation NAND. The neurons are connected by arrows. For any Boolean function $f$,
there exists a feed-forward A-type unorganised machine $A$ that `represents'
$f$. That is, there is always an A-type that given an input vector of Boolean
values $x$ will output the vector of Boolean values $f(x)$ (see
Section~\ref{sec:inputAndOutput}). Throughout this paper we use the term
`A-types' to refer to Turing's A-type unorganised machines. We also apply this
term when we discuss our interpretation of Turing's
A-type unorganised machines and those of other researchers; we hope that the meaning is clear from the context.

In~\cite{turing1948} Turing introduced three models of computation:
A-types, B-type unorganised machines, and P-type unorganised machines. In our
research we only use A-types. However, we mention these other two models to
explain their relevance to our research.

The second ANN that Turing introduced was a special kind of A-type, which he
called a \emph{B-type unorganized machine}. These networks are effectively
A-types the arrows of which can be switched on and off by changing the state of
particular nodes in the network. Turing constructed these switchable arrows
with a particular configuration of nodes and arrows.
In the late 1940's A-types would have had to have been
directly implemented in hardware; Turing's B-type unorganised
machines offer a means of effectively reconfiguring the topology of a network
without reconfiguring hardware. Today, ANNs are
often implemented in software that is several levels of abstraction above computer
hardware; however, there may be novel architectures for which the reconfigurable
architecture B-type unorganized machine is useful.

In~\cite{turing1948} Turing introduced P-type unorganised machines.
Unlike a B-type, a P-type is not a special case of an A-type (nor is a P-type
a generalisation of an A-type). Turing used P-types to investigate learning.
This pioneering work would now be classed as an investigation into reinforcement
learning. For further details see Copeland~\cite{copeland1996proudfoot}.


Artificial neural networks have found wide application and are an active area of
research, yet only a few researchers have continued
Turing's work on A-types. In 1996 Copeland and
Proudfoot~\cite{copeland1996proudfoot} re-examined this research. The most
notable continuation of research into Turing's networks was conducted in 2001
by Teuscher~\cite{teuscherOne}. Teuscher experimented with A-types with fixed
input states; for instance, he used A-types in this manner to solve basic
pattern classification tasks and he showed that their dynamics are analogous to
those of a non-linear oscillator~\cite{teuscher2001sot}. Teuscher employed EAs to train
Turing's networks: he used linear data structures (linear chromosomes) to
represent B-types. Teuscher used
B-types with lists that prescribed whether each arrow in a B-type was in a
`connected' or `disconnected' state~\cite[p88]{teuscherOne}. These lists give
linear chromosomes for Teuscher's B-types.

Today, Turing's A-types can be considered a special class of Random
Boolean Networks~\cite[p25]{teuscherOne}. Random Boolean Networks are simple
discrete dynamical systems that are capable of complex behaviour; consequently,
they are useful for modelling complex systems such as gene regulation
mechanisms in biology and the
internet~\cite{kauffman2000investigations},~\cite{rohlf2009self}. Teuscher
investigated the non-linear dynamics of A-types~\cite[ch
5]{teuscherOne},~\cite{teuscher2001sot}. Recently, Bull~\cite{bull2009dynamical},
and Bull and Preene~\cite{bull5481dynamical} investigated the evolution of
A-type machines, and they considered this in the context of discrete dynamical
systems.

\section{Our Interpretation of Turing's A-types}\label{sec:ours}

In this section we present our definition of an A-type. This is an
interpretation of Turing's A-type unorganised machines, and has been influenced
by Teuscher's research~\cite{teuscherOne}. We also provide illustrations of our
A-types, and we compare our definition with those of Turing and Teuscher.

\subsection{Our Definition}

An A-type is a discrete, recurrent, synchronously updated ANN. The firing rule
for every neuron in an A-type is invoked simultaneously---we can imagine that all
neurons are updated via the same clock. Each of the instants at which the neurons in an
A-type are synchronously updated is called a moment.

In order to define A-types, we need the notion of an \emph{A-type graph}.
An A-type graph is a directed graph\footnote{When we talk of
a directed graph we allow multiple arrows in one or both directions between a given pair of nodes and we
allow loops (arrows with the same source and target nodes). Some authors
call this a directed multigraph.} with the following properties. Every node
has an indegree no greater than two. An A-type graph has a non-empty set of
nodes called {\em input nodes}, each of which has indegree zero. An A-type graph has a non-empty set of
nodes called {\em output nodes}, each of which has outdegree zero. The set of input nodes and set of output
nodes do not intersect. Arrows from an input node to an output node are not
permitted. Nodes that are not output nodes have no restriction on
their outdegree.

An A-type consists of an A-type graph and a
non-negative integer $\delta$, called the \emph{delay time}. We interpret the
nodes of the graph as the neurons of an ANN, and the arrows of the graph as the interconnections. The delay time determines the number of moments from when
information first enters the input nodes to when we start to
collect information from the output nodes (we elaborate on this in
Section~\ref{sec:inputAndOutput}). We call the number of input (output) nodes
of an A-type its \emph{input (output) dimension}. Because A-types are recurrent, A-type graphs
can have closed paths.

An A-type is a Boolean ANN. Consider an A-type $A$. Each interconnection of $A$
carries exactly one bit of information per moment. That is, we associate a
Boolean variable with every arrow in the A-type graph of $A$. Every node in $A$
has a firing rule that is a Boolean function (of the variables entering that node).
Furthermore, every node in an A-type has a Boolean variable associated with
it. We call this variable the \emph{state} of that node.  In general, the state of a node
varies from moment to moment.  At any moment the
output of a node is equal to the state of the node.

We classify every node that is not an input node into one of two types depending
on its firing rule: nand nodes and delay nodes. A nand node $q$ has an indegree of two and its
firing rule is NAND. That is, let $a$ and $b$ denote the Boolean values
associated with the respective arrows entering $q$ at moment $t$. At moment $(t
+ 1)$ the state of $q$ is $a$ NAND $b$. A delay node $d$ has an indegree of one
and its firing rule is the identity. That is, let $a$ denote the Boolean value
associated with the arrow entering $d$ at moment $t$. At moment $(t + 1)$ the
state of $d$ is $a$. A nand node can accept two inputs from a single nand or delay node. Note that we initialize the state of every non-input node
to zero.  We explain the rules for initialising and updating input nodes in Section~\ref{sec:processingInfo}.

\subsection{Illustrations}

\begin{figure}[htp]
\centering
\begin{pspicture}[showgrid=false](3,2)
\psscalebox{.7}{
%
\rput(0,2.5){\circlenode{A}{\phantom{0}}}
\rput(0,0.5){\circlenode{B}{\phantom{0}}}
\rput(2,2.5){\trinode[trimode=R]{C}{\phantom{0}}}
\rput(2,0.5){\circlenode[trimode=R]{D}{\phantom{0}}}
\rput(4,1.5){\circlenode[doubleline=true]{E}{\phantom{0}}}
\ncline{->}{A}{C}
\ncline{->}{A}{D}
\ncline{->}{B}{D}
\ncline{->}{C}{E}
\ncline{->}{D}{E}
}
\end{pspicture}
\caption{An A-type graph.}\label{fig:joinNodes}
\end{figure}
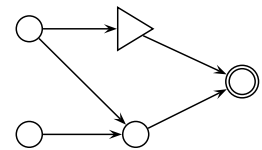

In graph theory diagrams are employed to represent a graph. We use
similar diagrams for our A-type graphs. Input nodes are represented by
circles with no incoming arrows. Nand nodes are represented by circles that have
two incoming arrows. Delay nodes are represented by triangles that have one
incoming arrow. Output nodes are denoted by doubled circles or doubled
triangles. We illustrate these conventions in Figure~\ref{fig:nodes}. In
Figure~\ref{fig:joinNodes} we depict a simple A-type.

\begin{figure}[htp]
\centering
\subfigure[An input node.]
{
\label{fig:nodes_a}
\begin{pspicture}[showgrid=false](3,2)
\psscalebox{.7}
{
%
\rput(1,1.5){\circlenode{C}{\phantom{X}}}
\rput(3.5,2.5){\circlenode[linecolor=white]{D}{$y_1$}}
\rput(3.5,2.0){\circlenode[linecolor=white]{E}{$y_2$}}
\rput(3.5,1.5){\circlenode[linecolor=white]{F}{.}}
\rput(3.5,1.0){\circlenode[linecolor=white]{G}{.}}
\rput(3.5,0.5){\circlenode[linecolor=white]{H}{.}}
\rput(3.5,0.0){\circlenode[linecolor=white]{I}{$y_r$}}
\ncline{->}{C}{D}
\ncline{->}{C}{E}
\ncline{->}{C}{I}
}
\end{pspicture}
}
%
%
\subfigure[A nand node.]
{
\label{fig:nodes_b}
\begin{pspicture}[showgrid=false](3,2)
\psscalebox{.7}
{
%
\rput(0,2.5){\circlenode[linecolor=white]{A}{$a$}}
\rput(0,0.5){\circlenode[linecolor=white]{B}{$b$}}
\rput(2,1.5){\circlenode{C}{\phantom{X}}}
\rput(4.5,2.5){\circlenode[linecolor=white]{D}{$y_1$}}
\rput(4.5,2.0){\circlenode[linecolor=white]{E}{$y_2$}}
\rput(4.5,1.5){\circlenode[linecolor=white]{F}{.}}
\rput(4.5,1.0){\circlenode[linecolor=white]{G}{.}}
\rput(4.5,0.5){\circlenode[linecolor=white]{H}{.}}
\rput(4.5,0.0){\circlenode[linecolor=white]{I}{$y_r$}}
\ncline{->}{A}{C}
\ncline{->}{B}{C}
\ncline{->}{C}{D}
\ncline{->}{C}{E}
\ncline{->}{C}{I}
}
\end{pspicture}
}
%
%
\subfigure[A delay node.]
{
\label{fig:nodes_c}
\begin{pspicture}[showgrid=false](3.5,2)
\psscalebox{.7}
{
%
\rput(0,1.5){\circlenode[linecolor=white]{A}{$a$}}
\rput(2,1.5){\trinode[trimode=R]{C}{\phantom{0}}}
\rput(4.5,2.5){\circlenode[linecolor=white]{D}{$y_1$}}
\rput(4.5,2.0){\circlenode[linecolor=white]{E}{$y_2$}}
\rput(4.5,1.5){\circlenode[linecolor=white]{F}{.}}
\rput(4.5,1.0){\circlenode[linecolor=white]{G}{.}}
\rput(4.5,0.5){\circlenode[linecolor=white]{H}{.}}
\rput(4.5,0.0){\circlenode[linecolor=white]{I}{$y_r$}}
\ncline{->}{A}{C}
\ncline{->}{C}{D}
\ncline{->}{C}{E}
\ncline{->}{C}{I}
}
\end{pspicture}
}
%
%
\subfigure[Output nodes.]
{
\label{fig:nodes_d}
\begin{pspicture}[showgrid=false](4,2)
\psscalebox{.7}
{
%
%
\rput(0,2.5){\circlenode[linecolor=white]{A}{$a$}}
\rput(0,0.5){\circlenode[linecolor=white]{B}{$b$}}
\rput(2,1.5){\circlenode[doubleline=true]{C}{\phantom{X}}}
\ncline{->}{A}{C}
\ncline{->}{B}{C}
%
%
%
\rput(3,1.5){\circlenode[linecolor=white]{D}{a}}
\rput(5,1.5){\trinode[trimode=R,doubleline=true]{E}{\phantom{0}}}
\ncline{->}{D}{E}
}
\end{pspicture}
}
\caption{Illustrating the types of nodes in an A-type. Note
that for any particular moment the Boolean values $y_1,\ldots, y_r$ associated
to the arrows exiting a given node are
all identical.}\label{fig:nodes}
\end{figure}
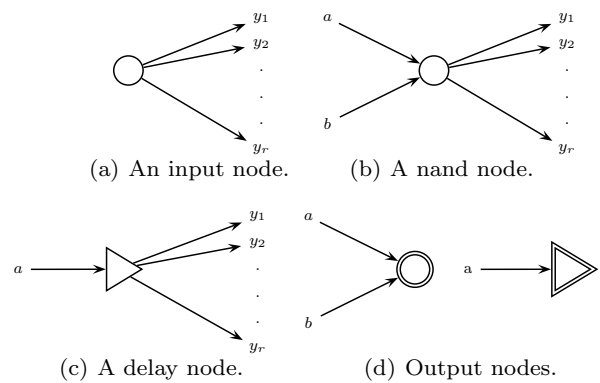

\subsection{Processing Information}\label{sec:processingInfo}

In this section we describe how we employ A-types to process information. By a
{\em Boolean vector} we mean a vector the components of which are all either
0 or 1.  We denote by $S_m$ the set of all $m$-component Boolean vectors.  We
now explain how A-types can accept and output sequences of Boolean vectors.

\subsubsection{Input and Output}\label{sec:inputAndOutput}

Consider an A-type $A$ that has input dimension $n$. To enable us to input
information into $A$ we adopt the following update rule for the input nodes of
$A$. Choose an ordering on the set of input nodes.
Suppose we are given a sequence of $n$-component Boolean vectors $(x_0,
\ldots , x_q)$. For the first $q$ moments, the states of the input
nodes of $A$ at moment $t$ are given by the components of $x_t$ in the appropriate order.
In particular, the initial states of the input nodes are determined by $x_0$.
 We say that at moment $t$,
$x_t$ is input into $A$. We adopt the convention that after the input vectors
are used up the states of the input nodes remain constant, keeping the values from the
final input vector. That is, for moments $t > q$ the states of the input nodes
are given by the components of $x_q$.

Consider an A-type $A$ that has delay $\delta$ and output dimension $p$.
We collect output information from
the output nodes of $A$, starting not at moment 0 but at moment $\delta$; the idea is that it may take
data some time to percolate through the A-type and reach the output nodes.
At each moment $t \geq \delta$ the
states of the output nodes of $A$
generate a $p$-component Boolean vector $y_t$. We say that at that moment $A$
outputs $y_t$.

A-types can be viewed as non-linear dynamical systems~\cite[p132]{teuscherOne}.
Because our A-types accept sequential data, they are analogous to
non-linear oscillators that are subjected to a driving force that is generally
not constant.  Note that when we use an A-type to process binary data, the delay $\delta$ is a parameter which is independent of the input data.  If an A-type is to represent a sequential function in the sense of Section~\ref{sec:repSeqFunctions} below, it must have the special property that the time for the input data to travel to the output(s) should not depend on the choice of input data.

\subsubsection{Clamped Input Mode}\label{sec:clampedAtypes}

Consider the special case when a single Boolean vector $x_0$ is
input into an A-type $A$. The states of the input nodes of $A$ stay constant,
with values determined by $x_0$. In
this case we say that the input nodes of $A$ are \emph{clamped} by $x_0$, and we
say that we are operating $A$ in {\em clamped mode}.

Consider an A-type $A$, with delay $\delta$. Let $A$ be clamped by some input vector
$x_0$. If the states of the
output nodes of $A$ are constant  for all moments $t \geq \delta$ then we say
 $A$ is \emph{clampable} with respect to the input $x_0$.  We say $A$ is {\em clampable}
 if it is clampable for every $x_0$.

We can operate an A-type in clamped input mode even when
it is not clampable. Because the graph of an A-type is finite, if an A-type is
operated in clamped mode then eventually the output becomes periodic. For a clampable
A-type, this period is always one.

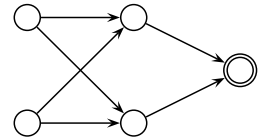
\begin{figure}[htp]
\centering
\begin{pspicture}[showgrid=false](3,2)
\psscalebox{.7}{
%
\rput(0,2.5){\circlenode{A}{\phantom{0}}}
\rput(0,0.5){\circlenode{B}{\phantom{0}}}
\rput(2,2.5){\circlenode{C}{\phantom{0}}}
\rput(2,0.5){\circlenode[trimode=R]{D}{\phantom{0}}}
\rput(4,1.5){\circlenode[doubleline=true]{E}{\phantom{0}}}
\ncline{->}{A}{C}
\ncline{->}{A}{D}
\ncline{->}{B}{C}
\ncline{->}{B}{D}
\ncline{->}{C}{E}
\ncline{->}{D}{E}
}
\end{pspicture}
\caption{A clampable A-type with delay $\delta = 2$.}\label{fig:logicalAnd}
\end{figure}

We now present an example of a clampable A-type. Consider the A-type
$A_{\wedge}$, with a delay of $\delta = 2$, shown in
Figure~\ref{fig:logicalAnd}.  It is easy to check that $A_{\wedge}$ is clampable and
that for every input $[a,b]\footnote{For typographical reasons, we write a
Boolean vector in row vector form in the text and in column vector form in our
figures.}\in S_2$, the eventual output is $a\wedge b$.

A Boolean function is a function from $S_n$ to $S_p$ for some positive
integers $n$ and $p$. Consider a Boolean function $f$ and a clampable A-type
$A$. We say that $A$ \emph{represents} the Boolean function $f$ if the following
 holds: for each $x\in S_n$, if $A$
is clamped with respect to the input $x$ then $A$ outputs the constant sequence of vectors
$f(x)$. For example, Logical AND $\wedge$ maps $S_2$ to $S_1$ and it is clear
from the discussion in the previous paragraph that
the A-type $A_{\wedge}$ shown in Figure~\ref{fig:logicalAnd} represents
$\wedge$. It can be shown that for any Boolean function $f$, there exists a feed-forward
A-type without delay nodes that represents $f$; Figure~\ref{figure:needDelayMachinesOne_a} illustrates how to construct an A-type to represent a Boolean function which is built from $\vee$ and $\wedge$.  In Section~\ref{sec:repSeqFunctions} we generalize
our definition of what it means for an A-type to represent a function.

\subsubsection{Sequential Input Mode}

In Section~\ref{sec:inputAndOutput} we defined a way of inputting information
into an A-type so that A-types can accept and return sequences of Boolean
vectors. We considered constant input and output sequences in Section~\ref{sec:clampedAtypes}. In general, the
sequences that A-types accept need not be constant.

Consider an A-type $A$ with delay $\delta$, input dimension $n$ and
output dimension $p$. Recall that at every moment $t$, $A$ accepts an
$n$-component Boolean vector $x_t$, and for each moment $t \geq \delta$, $A$ returns a
$p$-component Boolean vector $y_t$.  We say that we are operating $A$ in {\em sequential mode}.

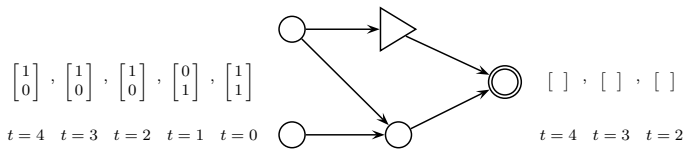
\begin{figure}[htp]
\centering
\begin{pspicture}[showgrid=false](8.5,2)
\psscalebox{.7}{
%
%
\rput(0.0,1.5){$\twoVector{1}{0}$}
\rput(0.5,1.5){,}
\rput(1.0,1.5){$\twoVector{1}{0}$}
\rput(1.5,1.5){,}
\rput(2.0,1.5){$\twoVector{1}{0}$}
\rput(2.5,1.5){,}
\rput(3.0,1.5){$\twoVector{0}{1}$}
\rput(3.5,1.5){,}
\rput(4.0,1.5){$\twoVector{1}{1}$}
%
\rput(5,2.5){\circlenode{A}{\phantom{0}}}
\rput(5,0.5){\circlenode{B}{\phantom{0}}}
\rput(7,2.5){\trinode[trimode=R]{C}{\phantom{0}}}
\rput(7,0.5){\circlenode[trimode=R]{D}{\phantom{0}}}
\rput(9,1.5){\circlenode[doubleline=true]{E}{\phantom{0}}}
\ncline{->}{A}{C}
\ncline{->}{A}{D}
\ncline{->}{B}{D}
\ncline{->}{C}{E}
\ncline{->}{D}{E}
%
\rput(10,1.5){$\oneVector{\phantom{l}}$}
\rput(10.5,1.5){,}
\rput(11,1.5){$\oneVector{\phantom{l}}$}
\rput(11.5,1.5){,}
\rput(12,1.5){$\oneVector{\phantom{l}}$}
%
\rput(0.0,0.5){$t = 4$}
\rput(1.0,0.5){$t = 3$}
\rput(2.0,0.5){$t = 2$}
\rput(3.0,0.5){$t = 1$}
\rput(4.0,0.5){$t = 0$}
%
\rput(10,0.5){$t = 4$}
\rput(11,0.5){$t = 3$}
\rput(12,0.5){$t = 2$}
}
\end{pspicture}
\caption{A sequence of five input vectors, and an A-type with delay $\delta =
2$. Three output vectors are expected in response to
these input vectors.}\label{fig:computeZero}
\end{figure}

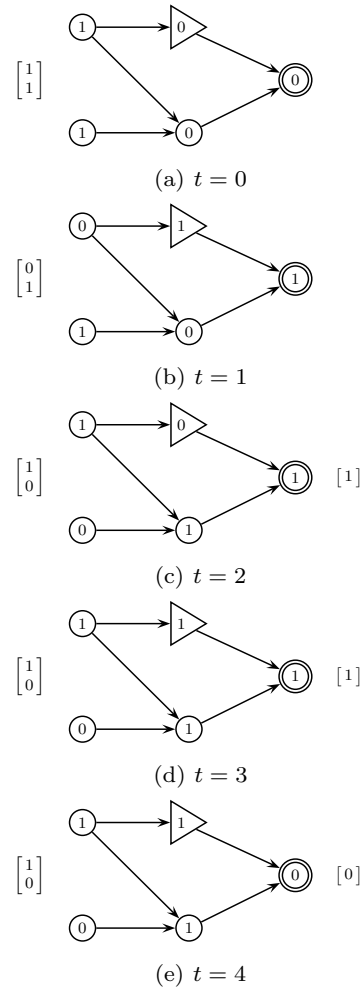
\begin{figure}[htp]
\centering
\subfigure[$t = 0$]
{
\label{fig:computeOne_a}
\begin{pspicture}[showgrid=false](4.5,2)
\psscalebox{.7}{
%
%
\rput(0,1.5){$\twoVector{1}{1}$}
%
\rput(1,2.5){\circlenode{A}{1}}
\rput(1,0.5){\circlenode{B}{1}}
\rput(3,2.5){\trinode[trimode=R]{C}{0}}
\rput(3,0.5){\circlenode[trimode=R]{D}{0}}
\rput(5,1.5){\circlenode[doubleline=true]{E}{0}}
\ncline{->}{A}{C}
\ncline{->}{A}{D}
\ncline{->}{B}{D}
\ncline{->}{C}{E}
\ncline{->}{D}{E}
%
%
}
\end{pspicture}
}
\subfigure[$t = 1$]
{
\label{fig:computeOne_b}
\begin{pspicture}[showgrid=false](4.5,2)
\psscalebox{.7}{
%
%
\rput(0,1.5){$\twoVector{0}{1}$}
%
\rput(1,2.5){\circlenode{A}{0}}
\rput(1,0.5){\circlenode{B}{1}}
\rput(3,2.5){\trinode[trimode=R]{C}{1}}
\rput(3,0.5){\circlenode[trimode=R]{D}{0}}
\rput(5,1.5){\circlenode[doubleline=true]{E}{1}}
\ncline{->}{A}{C}
\ncline{->}{A}{D}
\ncline{->}{B}{D}
\ncline{->}{C}{E}
\ncline{->}{D}{E}
%
%
}
\end{pspicture}
}
\subfigure[$t = 2$]
{
\label{fig:computeOne_c}
\begin{pspicture}[showgrid=false](4.5,2)
\psscalebox{.7}{
%
%
\rput(0,1.5){$\twoVector{1}{0}$}
%
\rput(1,2.5){\circlenode{A}{1}}
\rput(1,0.5){\circlenode{B}{0}}
\rput(3,2.5){\trinode[trimode=R]{C}{0}}
\rput(3,0.5){\circlenode[trimode=R]{D}{1}}
\rput(5,1.5){\circlenode[doubleline=true]{E}{1}}
\ncline{->}{A}{C}
\ncline{->}{A}{D}
\ncline{->}{B}{D}
\ncline{->}{C}{E}
\ncline{->}{D}{E}
%
\rput(6,1.5){$\oneVector{1}$}
}
\end{pspicture}
}
\subfigure[$t = 3$]
{
\label{fig:computeOne_d}
\begin{pspicture}[showgrid=false](4.5,2)
\psscalebox{.7}{
%
%
\rput(0,1.5){$\twoVector{1}{0}$}
%
\rput(1,2.5){\circlenode{A}{1}}
\rput(1,0.5){\circlenode{B}{0}}
\rput(3,2.5){\trinode[trimode=R]{C}{1}}
\rput(3,0.5){\circlenode[trimode=R]{D}{1}}
\rput(5,1.5){\circlenode[doubleline=true]{E}{1}}
\ncline{->}{A}{C}
\ncline{->}{A}{D}
\ncline{->}{B}{D}
\ncline{->}{C}{E}
\ncline{->}{D}{E}
%
\rput(6,1.5){$\oneVector{1}$}
}
\end{pspicture}
}
\subfigure[$t = 4$]
{
\label{fig:computeOne_e}
\begin{pspicture}[showgrid=false](4.5,2)
\psscalebox{.7}{
%
%
\rput(0,1.5){$\twoVector{1}{0}$}
%
\rput(1,2.5){\circlenode{A}{1}}
\rput(1,0.5){\circlenode{B}{0}}
\rput(3,2.5){\trinode[trimode=R]{C}{1}}
\rput(3,0.5){\circlenode[trimode=R]{D}{1}}
\rput(5,1.5){\circlenode[doubleline=true]{E}{0}}
\ncline{->}{A}{C}
\ncline{->}{A}{D}
\ncline{->}{B}{D}
\ncline{->}{C}{E}
\ncline{->}{D}{E}
%
\rput(6,1.5){$\oneVector{0}$}
}
\end{pspicture}
}
\caption{Snapshots of the A-type shown in Figure~\ref{fig:computeOne} over the
first five moments.  The numbers written inside the nodes are the states of
the nodes at the given moment.}\label{fig:computeOne}
\end{figure}

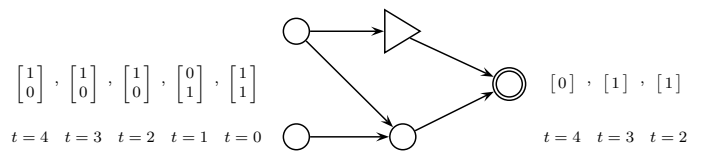
\begin{figure}[htp]
\centering
\begin{pspicture}[showgrid=false](8.5,2)
\psscalebox{.7}{
%
%
\rput(0.0,1.5){$\twoVector{1}{0}$}
\rput(0.5,1.5){,}
\rput(1.0,1.5){$\twoVector{1}{0}$}
\rput(1.5,1.5){,}
\rput(2.0,1.5){$\twoVector{1}{0}$}
\rput(2.5,1.5){,}
\rput(3.0,1.5){$\twoVector{0}{1}$}
\rput(3.5,1.5){,}
\rput(4.0,1.5){$\twoVector{1}{1}$}
%
\rput(5,2.5){\circlenode{A}{\phantom{0}}}
\rput(5,0.5){\circlenode{B}{\phantom{0}}}
\rput(7,2.5){\trinode[trimode=R]{C}{\phantom{0}}}
\rput(7,0.5){\circlenode[trimode=R]{D}{\phantom{0}}}
\rput(9,1.5){\circlenode[doubleline=true]{E}{\phantom{0}}}
\ncline{->}{A}{C}
\ncline{->}{A}{D}
\ncline{->}{B}{D}
\ncline{->}{C}{E}
\ncline{->}{D}{E}
%
\rput(10,1.5){$\oneVector{0}$}
\rput(10.5,1.5){,}
\rput(11,1.5){$\oneVector{1}$}
\rput(11.5,1.5){,}
\rput(12,1.5){$\oneVector{1}$}
%
\rput(0.0,0.5){$t = 4$}
\rput(1.0,0.5){$t = 3$}
\rput(2.0,0.5){$t = 2$}
\rput(3.0,0.5){$t = 1$}
\rput(4.0,0.5){$t = 0$}
%
\rput(10,0.5){$t = 4$}
\rput(11,0.5){$t = 3$}
\rput(12,0.5){$t = 2$}
}
\end{pspicture}
\caption{From Figure ~\ref{fig:computeOne} we can determine the response of the
A-type shown in Figure~\ref{fig:computeZero}. Here we show the full
sequence of input and output vectors for this A-type.}\label{fig:computeTwo}
\end{figure}

In Figure~\ref{fig:computeZero} we illustrate a simple A-type $A$ with delay
$\delta = 2$, and an input sequence of 5 Boolean vectors. The A-type $A$ returns an
output sequence consisting of 3 Boolean vectors. In Figure~\ref{fig:computeOne} we illustrate how
$A$ changes over these moments. Each subfigure is a
snapshot of the entire A-type at a particular moment. We
give the input Boolean vector, the states of the nodes of $A$ and the output vector at that moment. In
Figure~\ref{fig:computeTwo} we illustrate $A$, the input sequence for the first
five moments, and the output sequence for the first five moments\footnote{Note
that the rightmost entry of an input sequence
is the input vector for the earliest moment.}.

Note that (by our convention introduced in Section~\ref{sec:inputAndOutput}) if
a sequence of $l$ Boolean vectors is input into an A-type then for every moment
$t > l$ the states of the input nodes of that A-type are constant. This
convention serves to `shunt' information through an A-type.
For instance, in Figure~\ref{fig:computeZero} the initial input sequence and
the desired output sequence have length 3, but 5 moments are needed to collect
the output because $\delta= 2$.  Our shunting convention ensures that the input
states are well-defined for the final two moments.

%
%

\subsection{Representing Sequential Functions} \label{sec:repSeqFunctions}

Here we explain how to associate a function to an A-type. In
Section~\ref{sec:clampedAtypes} we defined what it means for a clampable A-type
operating in clamped input mode to represent a
Boolean function. Here we generalize this notion.

Let $S_{m,l}$ denote the set of all sequences of length $l$ consisting of
$m$-component Boolean vectors. Note that $S_{m,1}= S_m$.  Consider a function
$f$ from $S_{n,k}$ to
$S_{p,l}$, for some positive integers $k$, $l$, $n$, and $p$.  We call $f$ a
{\em sequential Boolean function}. A Boolean function $f$ is the special case of a
sequential Boolean function with $k = l = 1$. We say that an A-type $A$
\emph{represents} $f$ if for every $x \in S_{n,k}$, when $A$ accepts $x$
it outputs the sequence $f(x)$. So if $A$ represents $f$ then the input
dimension of $A$ must be $n$ and the output dimension of $A$ must be $p$.

For example, consider serial addition. We can describe this in terms of a
sequential function $f_+$ which maps $S_{2,l}$ to $S_{1,(l + 1)}$, for some
positive integer $l$. Given an input sequence $x$, the first entries of the
vectors in $x$ give the binary encoding for some integer $a$, the second entries
of the vectors in $x$ give the binary encoding for some integer $b$, and the
 entries of the vectors in $f(x)$ give the binary encoding for the integer
$(a + b)$. In Section~\ref{sec:columnwiseFunctions} we describe another class of
sequential functions, the columnwise Boolean functions.


Let us touch upon the possible functions that our A-types can represent when
operated in sequential input mode. Recall from Section~\ref{sec:clampedAtypes} that
any clamped Boolean function can be represented by a clampable feed-forward
A-type. We can regard representing functions in clamped mode as a special case of representing
functions in sequential mode: the input sequences have length 1 and the output
sequences are required to be constant.
 Because of this, sequential tasks are
generally more difficult than clamped tasks.  In principle, one can devise an A-type that represents
binary addition of strings $s_1$ and $s_2$ or arbitrary length; however, in practice this is not
trivial\footnote{For example, Minsky~\cite[p27]{minsky1967cfa} describes a
McCulloch-Pitts network that performs serial addition. In principle, these
details could be used to construct an A-type that represents binary addition of
$s_1$ and $s_2$.}. It is impossible~\cite[p27]{minsky1967cfa} to devise an
A-type that represents binary multiplication
of strings $s_1$ and $s_2$ of arbitrary length.


\subsubsection{Columnwise Boolean Functions}\label{sec:columnwiseFunctions}

We define a \emph{columnwise Boolean function} as follows. Let $n,p$ be positive
integers and suppose we are given a positive integer $k$. For any Boolean function $f$ we
define columnwise $f$ to be the function that maps $S_{n,k}$ to $S_{p,k}$ for any
positive integer $k$, such that if $x_i$ denotes the $i$th term of an input
sequence and $y_i$ denotes the $i$th term of the corresponding output
sequence then $y_i = f(x_i)$.  This says that bits of the input in different columns
do not interact with each other.  Conversely, if $g$ is a columnwise Boolean function then
we call the underlying Boolean function clamped $g$.

\begin{figure}
\begin{center}
\begin{pspicture}[showgrid=false](4,4)
\psscalebox{.7}{
\rput(0,5){\circlenode{A}{\phantom{0}}}
\rput(0,3){\circlenode{B}{\phantom{0}}}
\rput(2,5){\trinode[trimode=R]{C}{\phantom{0}}}
\rput(2,3){\trinode[trimode=R]{D}{\phantom{0}}}
\rput(2,1){\circlenode{E}{\phantom{0}}}
\rput(4,5){\circlenode{F}{\phantom{0}}}
\rput(4,3){\circlenode{G}{\phantom{0}}}
\rput(6,4){\circlenode[doubleline=true]{H}{\phantom{0}}}
\nccurve[angleA=-135,angleB=225]{->}{A}{E}
\nccurve[angleA=-80,angleB=135]{->}{B}{E}
\nccurve[angleA=45,angleB=-135]{->}{E}{F}
\nccurve[angleA=0,angleB=-135]{->}{E}{G}
\ncline{->}{A}{C}
\ncline{->}{B}{D}
\ncline{->}{C}{F}
\ncline{->}{D}{G}
\ncline{->}{F}{H}
\ncline{->}{G}{H}
}
\end{pspicture}
\end{center}
\caption{An A-type with delay time $\delta = 3$ that represents columnwise
Exclusive OR.}
\label{fig:XOR}
\end{figure}
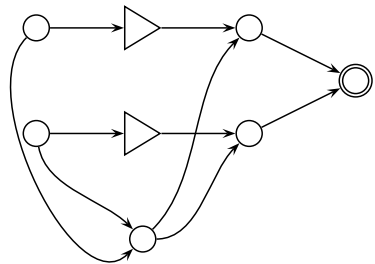

For example, let us consider columnwise Exclusive OR. The Boolean
function Exclusive OR $\oplus$ maps $S_2$ to $S_1$.  Columnwise Exclusive OR maps a sequence
$x$ of $k$ 2-component Boolean vectors to a sequence $y$ of $k$ 1-component
Boolean vectors, such that the $i$th term of $y$ is $\oplus(x_i)$, where $x_i$
denotes the $i$th term of $x$. It is easy to check that the A-type shown in
Figure~\ref{fig:XOR} represents columnwise Exclusive OR.
%
%

One can show that A-type representing a columnwise Boolean function $f$ also represents clamped $f$,
but the converse is false in general for the reasons discussed at the start of
Section~\ref{sec:delaysNeeded}.
%

%
%
%

\subsection{The Necessity of Delay Nodes}\label{sec:delaysNeeded}

Operating A-types in sequential mode brings some new challenges.  Data
travels through the A-type from input nodes to output nodes along various paths.  If these paths have different lengths
then the arrival times are not synchronised.  In order to represent sequential functions, it is
useful---and, we believe, sometimes necessary---to have a way to stagger the data.  This is why we introduced
delay nodes, which do not appear in Turing's original notion of an A-type.

We collected evidence that delay nodes are necessary for A-types to perform certain
sequential tasks. In particular, via computer simulations we collected evidence
that supports the following claim.

\medskip
\noindent \emph{Claim:} There does not exist an A-type without delay nodes that
represents columnwise \xor.

\medskip
We employed a blind search for A-types representing columnwise
Exclusive OR. We repeatedly constructed a random A-type with delay nodes and
tested whether it represented columnwise \xor. Similarly, we repeatedly
constructed a random A-type without delay nodes and tested whether it
represented columnwise \xor. The size of each A-type was randomly chosen
from the interval $[8,40]$. A sequence of $10^{4}$ randomly generated $2 \times
1$ input vectors was used to test whether an A-type represented columnwise
\xor: if an A-type represented columnwise \xor\ for such an input
sequence then it was deemed to do so for all input sequences. The results of
these searches are presented in Table~\ref{table:resultsForSequentialXOR}. In
summary, we discovered many A-types with delay nodes that represented columnwise
\xor; however, we failed to find a single A-type without delay nodes
that represented columnwise \xor.


\begin{table}
\caption{The results of our blind searches for A-types that represent columnwise
\xor.}
\label{table:resultsForSequentialXOR}
\centering
\begin{tabular}{@{} p{6.3cm} l r@{}}
\toprule
\multicolumn{2}{c}{With delay nodes} \\
\midrule
number of attempts  & $1.6 \times 10^{10}$ \\
probability that a node is constructed as a delay node  & $20\%$ \\
number of solutions & $1342$ \\
\toprule
\multicolumn{2}{c}{Without delay nodes} \\
\midrule
number of attempts  & $1.6 \times 10^{10}$ \\
probability that a node is constructed as a delay node  & $0\%$ \\
number of solutions & $0$ \\
\bottomrule
\end{tabular}
\end{table}

\begin{figure}
\begin{center}
\begin{tabular}{c}
\subfigure[Composing an A-type ($\delta = 4$) to represent $A \oplus B = (A \vee
B)\wedge(A \barwedge B)$. Note that the subgraph generated by the node set $\{ 0,1,2,3,5 \}$ represents $(A \vee B)$.  Also, the subgraph generated by the node set $\{ 0,1,4 \}$ represents $A \barwedge B$. Furthermore, the subgraph generated by the node set $\{ 4,5,6,7 \}$ represents AND.] {
\label{figure:needDelayMachinesOne_a}
\begin{minipage}{0.4\textwidth}
\centering
\begin{pspicture}[showgrid=false](7,4)
\psscalebox{0.7}{
$
\rput(0,4){\circlenode{A}{0}}
\rput(0,2){\circlenode{B}{1}}
\rput(2,5){\circlenode{C}{2}}
\rput(2,3){\circlenode{D}{3}}
\rput(2,1){\circlenode{E}{4}}
\rput(4,4){\circlenode{F}{5}}
\rput(6,3){\circlenode{G}{6}}
\rput(8,3){\circlenode[doubleline=true]{H}{7}}
\ncline[offset=3pt]{->}{A}{C}
\naput[npos=-0.6]{A}
\ncline[offset=-3pt]{->}{A}{C}
\ncline{->}{A}{E}
\ncline[offset=3pt]{->}{B}{D}
\naput[npos=-0.6]{B}
\ncline[offset=-3pt]{->}{B}{D}
\ncline{->}{B}{E}
\ncline{->}{C}{F}
\ncline{->}{D}{F}
\ncline{->}{E}{G}
\nbput[npos=0.1]{A \barwedge B}
\ncline{->}{F}{G}
\naput[npos=0.2]{A \vee B}
\ncline[offset=3pt]{->}{G}{H}
\naput[npos=1.6]{(A \vee B)\wedge(A \barwedge B)}
\ncline[offset=-3pt]{->}{G}{H}
$
}
\end{pspicture}
\\
\end{minipage}
}
\\
\subfigure[Inserting a delay machine into the A-type ($\delta = 4$) shown in
Figure~\ref{figure:needDelayMachinesOne_a}. This ensures that the two inputs into node $7$ are synchronized.] {
\label{figure:needDelayMachinesOne_b}
\begin{minipage}{0.4\textwidth}
\centering
\begin{pspicture}[showgrid=false](7,4)
\psscalebox{0.7}{
$
\rput(0,4){\circlenode{A}{0}}
\rput(0,2){\circlenode{B}{1}}
\rput(2,5){\circlenode{C}{2}}
\rput(2,3){\circlenode{D}{3}}
\rput(2,1){\circlenode{E}{4}}
\rput(4,4){\circlenode{F}{5}}
\rput(4,2){\trinode[trimode=R]{G}{\phantom{0}}}
\rput(6,3){\circlenode{H}{6}}
\rput(8,3){\circlenode[doubleline=true]{I}{7}}
\ncline[offset=3pt]{->}{A}{C}
\naput[npos=-0.6]{A}
\ncline[offset=-3pt]{->}{A}{C}
\ncline{->}{A}{E}
\ncline[offset=3pt]{->}{B}{D}
\naput[npos=-0.6]{B}
\ncline[offset=-3pt]{->}{B}{D}
\ncline{->}{B}{E}
\ncline{->}{C}{F}
\ncline{->}{D}{F}
\ncline{->}{E}{G}
\nbput[npos=0.1]{A \barwedge B }
\ncline{->}{F}{H}
\naput[npos=0.25]{A \vee B}
\ncline{->}{G}{H}
\nbput[npos=0.2]{A \barwedge B}
\ncline[offset=3pt]{->}{H}{I}
\naput[npos=1.6]{(A \vee B)\wedge(A \barwedge B)}
\ncline[offset=-3pt]{->}{H}{I}
$
}
\end{pspicture}
\\
\end{minipage}
}
\\
\subfigure[Generalising the solution shown in
Figure~\ref{figure:needDelayMachinesOne_b}. We insert two A-types into the
A-type shown in~\ref{figure:needDelayMachinesOne_a}. First, we insert an A-type
$I_{m}$ with a delay $\delta = m$, where $m$ is an odd positive
integer. Second, we insert an A-type $I_{m-1}$ with a delay $m-1$.
This ensures that the two inputs into node $6$ are synchronized. Consequently,
if we can discover an A-type without delay nodes that has an odd delay then we
can construct an A-type without delay nodes that represents columnwise \xor.] {
\label{figure:needDelayMachinesOne_c}
\begin{minipage}{0.4\textwidth}
\centering
\begin{pspicture}[showgrid=false](8.5,4)
\psscalebox{0.7}{
$
\rput(0,4){\circlenode{A}{0}}
\rput(0,2){\circlenode{B}{1}}
\rput(2,5){\circlenode{C}{2}}
\rput(2,3){\circlenode{D}{3}}
\rput(2,1){\circlenode{E}{4}}
\rput(4,4){\circlenode{F}{5}}
\rput[bl](4.5,1.5){\ovalnode{G}{\phantom{000}I_{m}\phantom{000}}}
\rput[bl](5.5,3.5){\ovalnode{H}{I_{m-1}}}
\rput(8.5,3){\circlenode{I}{6}}
\rput(10.5,3){\circlenode[doubleline=true]{J}{7}}
\ncline[offset=3pt]{->}{A}{C}
\naput[npos=-0.6]{A}
\ncline[offset=-3pt]{->}{A}{C}
\ncline{->}{A}{E}
\ncline[offset=3pt]{->}{B}{D}
\naput[npos=-0.6]{B}
\ncline[offset=-3pt]{->}{B}{D}
\ncline{->}{B}{E}
\ncline{->}{C}{F}
\ncline{->}{D}{F}
\ncline{->}{E}{G}
\nbput[npos=0.1]{A \barwedge B }
\ncline{->}{F}{H}
\naput[npos=0.4]{A \vee B}
\ncline{->}{G}{I}
\ncline{->}{H}{I}
\ncline[offset=3pt]{->}{I}{J}
\naput[npos=1.6]{(A \vee B)\wedge(A \barwedge B)}
\ncline[offset=-3pt]{->}{I}{J}
$
}
\end{pspicture}
\\
\end{minipage}
}
\end{tabular}
\caption{Using an expression that involves AND $\wedge$, NAND $\barwedge$, and
\ior\ $\vee$ to generate an A-type that represents \xor\ $\oplus$.}
\label{figure:needDelayMachinesOne}
\end{center}
\end{figure}
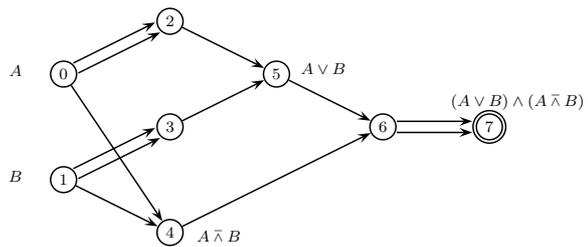
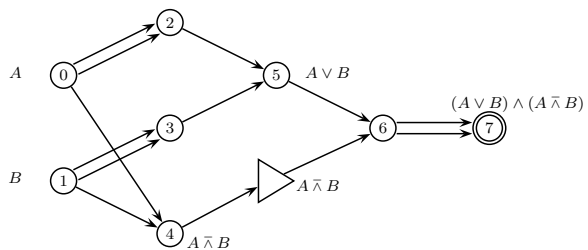
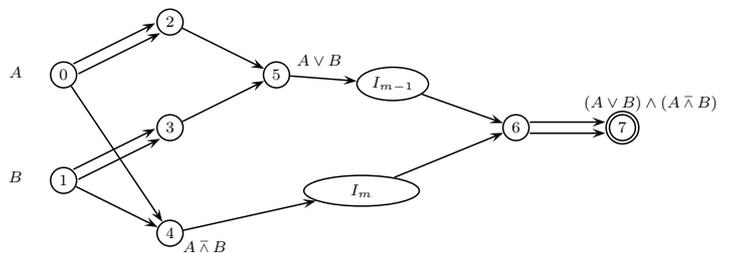

It is often the case that an A-type representing a clamped Boolean function can be modified to represent the corresponding columnwise Boolean function by adding some delay nodes.
The delay nodes are used to stagger data flowing through parts of an A-type and ensure the overall flow is synchronised. For example, consider again the Boolean function \xor.  We can write $A \oplus B$ as
$(A \vee B)\wedge(A \barwedge B)$. From this expression we devise a way to
construct an A-type $A_0$ that represents clamped \xor, using A-types
that represent columnwise \ior\ and columnwise AND; we illustrate this in
Figure~\ref{figure:needDelayMachinesOne_a}. Next we construct an A-type $A_1$ by
inserting a delay node into $A_0$; we illustrate this in
Figure~\ref{figure:needDelayMachinesOne_b}. This delay node
ensures that data is synchronised as it flows through $A_1$; consequently,
$A_0$ represents columnwise \xor.

Can one mimic the effect of the delay node using only nand nodes? We can formulate this question in terms of A-types that represent the identity.  Suppose there exists an A-type $I_m$ with a delay $\delta = m$, where $m$ is an odd positive integer, such that $I_m$ contains no delay nodes.  It is straightforward to find an A-type $I_{m-1}$ with even delay $m-1$ representing the identity function such that $I_{m-1}$ contains no delay nodes: Figure~\ref{figure:actualIdentitySolns_a} gives an A-type that works for the special case of delay 2, and we can obtain any even delay by concatenating copies of this A-type. Let us construct
an A-type $A_2$ as follows: we insert $I_{m}$ between nodes $4$ and $6$, and we insert $I_{m-1}$ with
between nodes $5$ and $6$ (if $m-1 = 0$ then we just put a single arrow directly from node $5$ to node $6$). See Figure~\ref{figure:needDelayMachinesOne_b}. This ensures that the two
inputs into node $6$ are synchronized.

It is clear from the above discussion that if there exists an A-type $I_m$ as above then we can mimic the effect of a delay node using only nand nodes.  The converse is also true, since a delay node represents the identity function with delay 1.  This motivates the following claim.

\medskip
\noindent \emph{Claim:} There does not exist an A-type without delay nodes
and with an odd delay that represents columnwise identity.

\medskip
\noindent The construction illustrated in Figure~\ref{figure:needDelayMachinesOne_c} shows that if this claim is false then the previous claim is also false.

We
employed a blind search for a counter-example to the  claim. We repeatedly
constructed a random A-type without delay nodes and tested whether it
represented columnwise identity. The size of each A-type was randomly chosen
from the interval $[3,20]$. For each A-type a sequence of $10^{4}$ randomly generated $1
\times 1$ input vectors was used to test whether the A-type represented
columnwise identity: if an A-type represented columnwise identity for such a
sequence then it was deemed to do so for all input sequences.
The results of these searches are presented in
Table~\ref{table:resultsForOddDelay}. In summary, we discovered many A-types
with even delay that represented columnwise identity; however, we failed to
find a single A-type with an odd delay that represented columnwise identity.

\begin{table}
\caption{The results of our blind searches for columnwise identity.}
\label{table:resultsForOddDelay}
\centering
\begin{tabular}{@{}l r@{}}
\toprule
Solution A-type & frequency \\
\midrule
number of attempts & 34000000000 \\
number of solutions with even delay  & $971789859$ \\
number of solutions with odd delay  & $0$\\
\bottomrule
\end{tabular}
\end{table}

From the above results, we conjecture that A-types with delay nodes operated in sequential mode can
represent a more general class of function than A-types without delay nodes. We
found experimental evidence that supports our claims, but we were not able to discover a formal mathematical proof.  We leave this as an open problem.

It is clear from the above discussions that we can implement a delay of any length by concatenating the following: (a) a single delay node, and (b) an A-type with even delay and without delay nodes that represents the identity.  Hence only a small number of delay nodes is needed in any given A-type.

\subsection{Comparison with Previous Definitions}

Our definition of an A-type, given above, differs from those of Turing and
Teuscher. The differences are in our allocation of input and output nodes, and
our introduction of delay nodes.

Turing did not precisely prescribe how information could be input and
output for A-types. To address this issue Teuscher~\cite[p32]{teuscherOne}
defined A-types with input and output nodes. Essentially, we have adopted
Teuscher's conventions for input and output nodes.

We introduce delay nodes so to allow our A-types to process sequential input.
Neither Turing nor Teuscher make use of delay
nodes. However, Teuscher~\cite[p67]{teuscherOne} investigates sequential
tasks by in effect employing two clock speeds: one for the rate of
information input and output, and one for the rate of information flow between neurons. We
chose to introduce delay machines because they allow a straightforward way
to interpret Turing's A-types so that they can operate with a sequential input.
Furthermore, the training algorithms that we employ for our A-types are useful
in both the clamped and the sequential settings---see Section~\ref{sec:simulations}.

\section{Learning via Evolution}\label{sec:evolution}

We now turn to our central problem: how to find an A-type that represents a given function $f$.
We implemented a reinforcement learning technique involving an EA which searches for `suitably
small' A-types that represent $f$.

In his pioneering paper \cite{turing1948} Turing examines reinforcement learning.  For instance, he defines a P-type machine to elaborate on some of his ideas.  Furthermore, Turing briefly mentions a `genetical search', but does not provide details of such a training method. One popular modern reinforcement learning
technique is EAs, and now their use to train ANNs is
established~\cite{floreano2008neuroevolution}. Teuscher used EAs to train
B-types~\cite{teuscherOne}. We also use EAs to train A-types.

In this section we outline a simple EA that we employ, and we present our
mutation and crossover operators. In
particular, we describe our efforts to devise useful crossover operators (see
Section~\ref{sec:evoOps}); further details can be found
in~\cite{orr2010evolving}. In Section~\ref{sec:simulations} we explain how we
tested our EA and we present the results of these tests.  Our EA works for
A-types in both clamped and sequential modes.

When we implement our EA, we need to assign values to various parameters.  Some of these values are task-dependent.  We give the parameter values in Section~\ref{sec:simulations}.

\subsection{Introducing our EA}

In Table~\ref{table:gaOne_outline} we give an outline of our EA. We call this EA
\gaOne. This EA is a straightforward implementation; for example, it is similar
to the scheme outlines in~\cite[ch 2]{eiben2003iec}, and the scheme outlined
in~\cite[ch 9]{mitchellOne}. Note that \gaOne\ is a steady state EA in that its
population has only a small variation from generation to generation. In later
sections we require the listing in
Table~\ref{table:gaOne_outline} to describe two special cases of \gaOne.

\begin{table}
\caption{An outline of the EA, \gaOne, that we use in this paper.}
\label{table:gaOne_outline}
\centering
\begin{tabular}{c}
\toprule
\gaOne\\
\midrule
\begin{minipage}{\columnwidth}
\begin{enumerate}
\item \emph{Create initial population:} Randomly generate a specified number of
candidate solutions of size within a specified range.
\item \emph{Iterate through successive generations:} Repeat until either the
population contains a solution or a maximum number of attempts have been performed.
\begin{enumerate}
\item \emph{perform a set number of crossovers:} Repeat a set number of times.
	\begin{enumerate}
	\item \emph{parent selection:} Select a pair of candidate solutions
	as parents. The fitter a candidate solution, the greater the probability that
	it is chosen as a parent.
	\item \emph{crossover:} For each parent pair combine information from both
	parents to produce a new candidate solution, which is
	added to the population.
	\item \emph{survivor selection:} Select a member of the population and delete
	it from the population.
	\end{enumerate}
\item \emph{perform a set number of mutations:} Repeat a set number of times.
	\begin{enumerate}
	\item \emph{mutation:} Randomly select a member of the population, copy it,
	slightly modify the copy, and add the modified copy to the population.
	\item \emph{survivor selection} Select a member of the population and delete it
	from the population.\end{enumerate}
\end{enumerate}
\item Return the fittest candidate solution in the population. If there is more
than one candidate solution with the lowest fitness of the population then we
randomly select an element from the set of such individuals.
\end{enumerate}
\end{minipage}\\
\bottomrule
\end{tabular}
\end{table}

\subsection{Chromosomes}

Our candidate solutions are graph-like; this is made explicit by our use
of an A-type graph to define an A-type.  An A-type graph can be represented by
an adjacency list. Teuscher~\cite[p88]{teuscherOne} demonstrated that A-types
(and B-types) can be assigned linear chromosomes. If an EA employs linear
chromosomes then it is easy to implement simple crossover and mutation
operators; for example, bit-flipping mutation and one-point crossover~\cite[ch
3]{eiben2003iec}.

We choose to represent A-types with graph chromosomes because it allows a
straightforward implementation of some graph-theoretic manipulations on A-types. In particular,
adding and removing topologically connected subgraphs from the graph of an
A-type becomes straightforward. We encode an A-type graph as an object which has a
collection of node objects associated to it; each node object can reference other node
objects. This approach has two advantages: it captures the topology of an A-type
graph, and it does not impose an artificial ordering on the nodes.

\subsection{The Initial Population}\label{sec:init_pop}

Our EA requires an initial population of A-types to be created.  To do this, random A-types are generated with size between a specified upper bound and a specified lower bound.  (We define the \emph{size} of an A-type $A$ to be the number of nodes it contains, and we denote this by $|A|$.)  The mutation and crossover operators can change the size of A-types, so subsequent populations can contain A-types whose size is outside the original bounds.

\subsection{Evolutionary Operators}\label{sec:evoOps}

Our EA involves mutation and crossover operators.
Here we describe our implementation of these operators.

\subsubsection{Mutation}

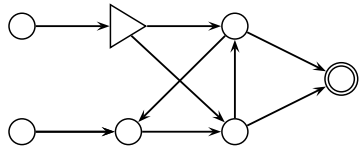
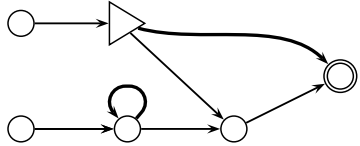
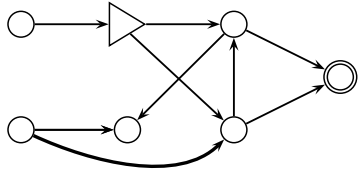
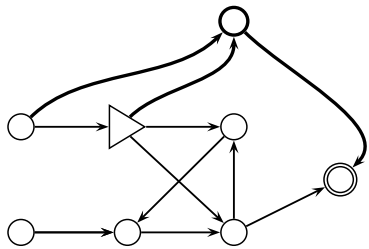
\begin{figure}[!ht]
\begin{center}
\subfigure[ The original A-type.]
{
\label{fig:mutant_a}
\begin{pspicture}[showgrid=false](5,3)
\psscalebox{.7}
{
\rput(0,3){\circlenode{A}{\phantom{0}}}
\rput(0,1){\circlenode{B}{\phantom{0}}}
\rput(2,3){\trinode[trimode=R]{C}{\phantom{0}}}
\rput(2,1){\circlenode{D}{\phantom{0}}}
\rput(4,3){\circlenode{E}{\phantom{0}}}
\rput(4,1){\circlenode{F}{\phantom{0}}}
\rput(6,2){\circlenode[doubleline=true]{G}{\phantom{0}}}
\ncline[linewidth=1.0pt]{->}{A}{C}
\ncline[linewidth=1.0pt]{->}{B}{D}
\ncline[linewidth=1.0pt]{->}{B}{D}
\ncline[linewidth=1.0pt]{->}{C}{E}
\ncline[linewidth=1.0pt]{->}{C}{F}
\ncline[linewidth=1.0pt]{->}{D}{F}
\ncline[linewidth=1.0pt]{->}{E}{G}
\ncline[linewidth=1.0pt]{->}{E}{D}
\ncline[linewidth=1.0pt]{->}{F}{E}
\ncline[linewidth=1.0pt]{->}{F}{G}
}
\end{pspicture}
}
\subfigure[A smaller mutant.]{
\label{fig:mutant_b}
\begin{pspicture}[showgrid=false](5,3)
\psscalebox{.7}
{
\rput(0,3){\circlenode{A}{\phantom{0}}}
\rput(0,1){\circlenode{B}{\phantom{0}}}
\rput(2,3){\trinode[trimode=R]{C}{\phantom{0}}}
\rput(2,1){\circlenode{D}{\phantom{0}}}
\rput(4,1){\circlenode{F}{\phantom{0}}}
\rput(6,2){\circlenode[doubleline=true]{G}{\phantom{0}}}
\ncline[linewidth=1.0pt]{->}{A}{C}
\ncline[linewidth=1.0pt]{->}{B}{D}
\ncline[linewidth=1.0pt]{->}{B}{D}
\ncline[linewidth=1.0pt]{->}{C}{F}
\nccurve[angleA=-15,angleB=135,linewidth=1.8pt]{->}{C}{G} 
\ncdiag[angleA=50,angleB=130,arm=0.8,linearc=.3,linewidth=1.8pt]{->}{D}{D}
\ncline[linewidth=1.0pt]{->}{D}{F}
\ncline[linewidth=1.0pt]{->}{F}{G}
}
\end{pspicture}
}
\subfigure[A mutant of the same size.]{
\label{fig:mutant_c}
\begin{pspicture}[showgrid=false](5,4)
\psscalebox{.7}
{
\rput(0,4){\circlenode{A}{\phantom{0}}}
\rput(0,2){\circlenode{B}{\phantom{0}}}
\rput(2,4){\trinode[trimode=R]{C}{\phantom{0}}}
\rput(2,2){\circlenode{D}{\phantom{0}}}
\rput(4,4){\circlenode{E}{\phantom{0}}}
\rput(4,2){\circlenode{F}{\phantom{0}}}
\rput(6,3){\circlenode[doubleline=true]{G}{\phantom{0}}}
\ncline[linewidth=1.0pt]{->}{A}{C}
\ncline[linewidth=1.0pt]{->}{B}{D}
\ncline[linewidth=1.0pt]{->}{B}{D}
\nccurve[angleA=-25,angleB=-135,linewidth=1.8pt]{->}{B}{F} 
\ncline[linewidth=1.0pt]{->}{C}{E}
\ncline[linewidth=1.0pt]{->}{C}{F}
\ncline[linewidth=1.0pt]{->}{E}{G}
\ncline[linewidth=1.0pt]{->}{E}{D}
\ncline[linewidth=1.0pt]{->}{F}{E}
\ncline[linewidth=1.0pt]{->}{F}{G}
}
\end{pspicture}
}
\subfigure[A larger mutant.]{
\label{fig:mutant_d}
\begin{pspicture}[showgrid=false](5,4)
\psscalebox{.7}
{
\rput(0,3){\circlenode{A}{\phantom{0}}}
\rput(0,1){\circlenode{B}{\phantom{0}}}
\rput(2,3){\trinode[trimode=R]{C}{\phantom{0}}}
\rput(2,1){\circlenode{D}{\phantom{0}}}
\rput(4,3){\circlenode{E}{\phantom{0}}}
\rput(4,1){\circlenode{F}{\phantom{0}}}
\rput(6,2){\circlenode[doubleline=true]{G}{\phantom{0}}}
\rput(4,5){\circlenode[linewidth=1.8pt]{H}{\phantom{0}}} 
\ncline[linewidth=1.0pt]{->}{A}{C}
\nccurve[angleA=45,angleB=-135,linewidth=1.8pt]{->}{A}{H} 
\ncline[linewidth=1.0pt]{->}{B}{D}
\ncline[linewidth=1.0pt]{->}{B}{D}
\ncline[linewidth=1.0pt]{->}{C}{E}
\ncline[linewidth=1.0pt]{->}{C}{F}
\nccurve[angleA=45,angleB=-90,linewidth=1.8pt]{->}{C}{H} 
\ncline[linewidth=1.0pt]{->}{D}{F}
\ncline[linewidth=1.0pt]{->}{E}{D}
\ncline[linewidth=1.0pt]{->}{F}{E}
\ncline[linewidth=1.0pt]{->}{F}{G}
\nccurve[angleA=-45,angleB=45,linewidth=1.8pt]{->}{H}{G} 
}
\end{pspicture}
}
\caption{Three examples of mutation.}\label{fig:mutation}
\end{center}
\end{figure}

Our mutation operator manipulates an A-type graph. The
search space of our EA contains A-types that have a range of sizes.
Consequently, we construct a mutation operator that can alter the size of an A-type. More
precisely, our mutation operators accept an A-type $A_{in}$ and return an
A-type $A_{out}$ such that either $|A_{out}| = |A_{in}| - 1$, or  $|A_{out}| =
|A_{in}|$, or  $|A_{out}| = |A_{in}| + 1$. We achieve this by copying the input
A-type: $A_{out} \leftarrow A_{in}$, and performing one of the three following
operations. One, a node $n$ is removed (if possible) from
$A_{out}$ and there is a slight re-arrangement of the graph of $A_{out}$
in order to make $A_{out}$ into a valid A-type. Two, a single arrow is removed from
$A_{out}$ and a new arrow inserted in order to make $A_{out}$ into a valid A-type.
Three, a node $n$ is added to $A_{out}$ and arrows are added, and there is a slight re-arrangement of
the graph of $A_{out}$ in order to ensure that $n$ has an output arrow and
$A_{out}$ is a valid A-type. We illustrate these operations in Figure~\ref{fig:mutation}.

\subsubsection{Crossover}

Our crossover operator involves operations that respect the topology of the
graphs of the parent A-types.  The operator exchanges subgraphs of the parents.  Only
topologically connected chunks of the parent graphs are exchanged, and reconnection of
exchanged chunks involves only the insertion of arrows that bridge the
`boundaries' of these chunks. We make this more precise in the following
subsections. In this section we present a crossover scheme which employs these
ideas. In Section~\ref{sec:simulations} we describe our tests of this crossover
operator.

Our crossover operator accepts two parent
A-types \Venus , \Mars\ and returns one child A-type $C$. Two aspects of this crossover operator
require further explanation: the acceptor and donor subgraphs are graphs of a
particular type, and there are restrictions on the arrows that may be inserted
to reconnect the child $C$. We elaborate on these two aspects next.

\begin{table}
\caption{Our A-type crossover operator.}
\label{tab:crossover}
\begin{tabular}{c}
\toprule
Crossover \\
\midrule
\begin{minipage}{\columnwidth}
\begin{enumerate}
\item The child $C$ is assigned simply to be a copy
of the parent \Venus; that is, $C \leftarrow$ \Venus.
\item A subgraph of $C$ is chosen; we call this the {\em acceptor} \texttt{A}.
\item A subgraph of \Mars\
is chosen; we call this the {\em donor} \texttt{D}.
\item The subgraph \texttt{A} is removed from $C$ (any
arrows bridging $(C -$ \texttt{A} $)$ and \texttt{A} are also removed) and a
copy of \texttt{D} is inserted into $C$.
\item Arrows are added to $C$ so that $C$ is a valid
A-type:
\begin{enumerate}
\item Inserting arrows from $C -$ \texttt{A} to \texttt{D}. For each arrow
the source is randomly selected from the distal boundary of \texttt{A}.
\item Inserting arrows from \texttt{D} to $C -$ \texttt{A}. For each arrow
the source is randomly selected from the proximal boundary of \texttt{D}.
\end{enumerate}
\end{enumerate}
\end{minipage}\\
\bottomrule
\end{tabular}
\end{table}

The donor and acceptor are subgraphs of a particular kind, which we call
{\em radial subgraphs}. Consider a graph $G$ and a node $c \subseteq G$, which we call
the centre ($c$ is chosen randomly in the crossover operator).  If possible, we construct a radial subgraph of $G$ with $N$ nodes
about the centre $c$ as follows. Construct a set $S$ which initially
 contains only $c$. Let $\bar{S}$ denote the set of all nodes that are
adjacent\footnote{Two nodes are adjacent if they are the endpoints of a
particular arrow. That is, two nodes are adjacent if there is an arrow
connecting them.} to nodes in $S$ but are not already in $S$. We randomly select
elements of $\bar{S}$ and transfer them to $S$ until $|S| = N$ or
$|\bar{S}| = 0$ (where $|S|$ denotes the size of $S$). We repeat the above process of constructing a set of nodes that
are adjacent to $S$ and selecting from that set until $|S| = N$ or $|\bar{S}| =
0$. At any point if $|S| = N$ then we use $S$ to generate a subgraph from
$G$. This subgraph is the desired radial subgraph.

For each of the acceptor and donor sets, the size $N$ is a randomly chosen integer between 1 and a fixed proportion of the total size of the parent graph.  The crossover algorithm always exchanges
`localized' and connected subgraphs of the graphs of the parents.

When our crossover reconnects subgraphs in the graph of the child, arrows may
only be inserted between boundaries of the acceptor and donor subgraphs. To explain this
process we define two types of boundaries: proximal boundaries and distal
boundaries. Consider a graph $G$ with a subgraph $S$. Also, let $G-S$ denote the
complement of $S$. The {\em proximal boundary} of $S$ is the set of nodes in $S$ that
are adjacent to nodes in $G-S$. The {\em distal boundary} of $S$ is the set of nodes
in $G-S$ that are adjacent to nodes in $S$. For our crossover operator, the
final step of constructing the child requires the insertion of arrows between
the complement of the acceptor and the donor. Arrows are only
inserted between nodes in the distal boundary of the acceptor and the proximal
boundary of the donor.

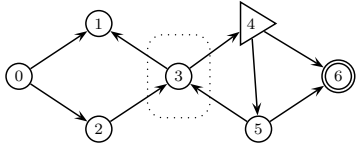
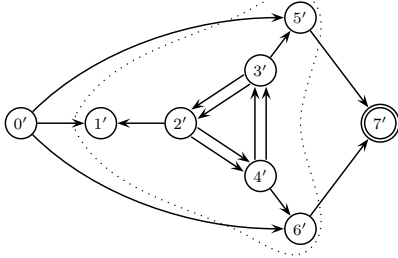
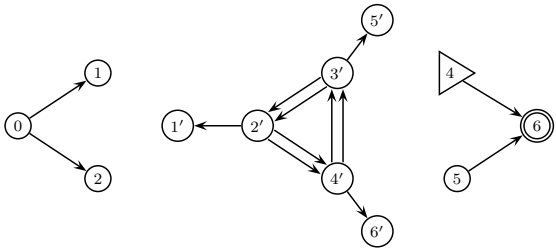
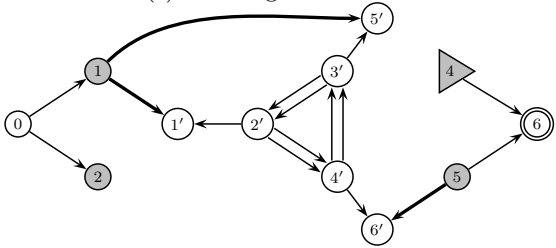
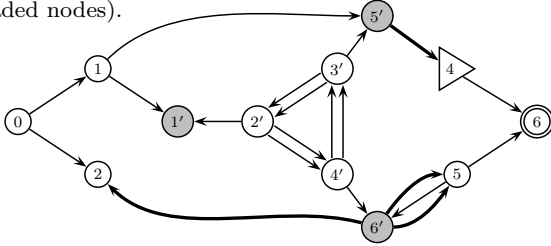
\begin{figure}[!ht]
\subfigure[The mother \Venus\ and its acceptor subgraph \texttt{A} $= \{ 3
\}$. Note that the proximal boundary = $\{ 3 \}$, and the distal boundary = $\{
1,2,4,5 \}$. ] {
\begin{minipage}{\linewidth}
\centering
\begin{pspicture}[showgrid=false](4.5,1.9)
\psscalebox{.7}{
\rput(0,1.5){\circlenode{A}{$0$}}
\rput(1.5,2.5){\circlenode{B}{$1$}}
\rput(1.5,0.5){\circlenode{C}{$2$}}
\rput(3,1.5){\circlenode{D}{$3$}}
\rput(4.5,2.5){\trinode[trimode=R]{E}{$4$}}
\rput(4.5,0.5){\circlenode{F}{$5$}}
\rput(6,1.5){\circlenode[doubleline=true]{G}{$6$}}
\ncline{->}{A}{B}
\ncline{->}{A}{C}
\ncline{->}{C}{D}
\ncline{->}{D}{B}
\ncline{->}{D}{E}
\ncline{->}{E}{F}
\ncline{->}{E}{G}
\ncline{->}{F}{D}
\ncline{->}{F}{G}
\psframe[framearc=0.5,linestyle=dotted](2.4,0.7)(3.6,2.3)
}
\end{pspicture}
\end{minipage}
}
%
%
\subfigure[The father \Mars\ and its donor subgraph \texttt{D} $= \{ 1' ,2' ,3'
,4' ,5' , 6' \}$. The proximal boundary = $\{ 1', 5' , 6' \}$, and the distal
boundary = $\{ 0' ,7' \}$. ] {
\begin{minipage}{\linewidth}
\centering
\begin{pspicture}[showgrid=false](5,3.3)
\psscalebox{.7}{
\rput(0,3){\circlenode{a}{$0'$}}
\rput(1.5,3){\circlenode{b}{$1'$}}
\rput(3,3){\circlenode{c}{$2'$}}
\rput(4.5,4){\circlenode{d}{$3'$}}
\rput(4.5,2){\circlenode{e}{$4'$}}
\rput(5.25,5){\circlenode{f}{$5'$}}
\rput(5.25,1){\circlenode{g}{$6'$}}
\rput(6.75,3){\circlenode[doubleline=true]{h}{$7'$}}
\ncline{->}{a}{b}
\nccurve[angleA=45,angleB=180]{->}{a}{f}
\nccurve[angleA=-45,angleB=180]{->}{a}{g}
\ncline{->}{c}{b}
\ncline[offset=-3pt]{->}{c}{e}
\ncline[offset=3pt]{->}{c}{e}
\ncline[offset=-3pt]{->}{d}{c}
\ncline[offset=3pt]{->}{d}{c}
\ncline{->}{d}{f}
\ncline[offset=-3pt]{->}{e}{d}
\ncline[offset=3pt]{->}{e}{d}
\ncline{->}{e}{g}
\ncline{->}{f}{h}
\ncline{->}{g}{h}
\psccurve[linestyle=dotted,showpoints=false]%
(1.05,2.8)(1.05,3.2)(4.0,5.0)(5.4,5.4)(5.25,3.0)(5.4,0.6)(4.0,1.0)
}
\end{pspicture}
\end{minipage}
}
%
%
%
\subfigure[Inserting \texttt{D} into \Venus\ $-$ \texttt{A} ]{
\begin{minipage}{\linewidth}
\centering
\begin{pspicture}[showgrid=false](7,3.6)
\psscalebox{.7}{
\rput(0,3){\circlenode{A}{$0$}}
\rput(1.5,4){\circlenode{B}{$1$}}
\rput(1.5,2){\circlenode{C}{$2$}}
\rput(3,3){\circlenode{b}{$1'$}}
\rput(4.5,3){\circlenode{c}{$2'$}}
\rput(6,4){\circlenode{d}{$3'$}}
\rput(6,2){\circlenode{e}{$4'$}}
\rput(6.75,5){\circlenode{f}{$5'$}}
\rput(6.75,1){\circlenode{g}{$6'$}}
\rput(8.25,4){\trinode[trimode=R]{E}{$4$}}
\rput(8.25,2){\circlenode{F}{$5$}}
\rput(9.75,3){\circlenode[doubleline=true]{G}{$6$}}
\ncline{->}{A}{B}
\ncline{->}{A}{C}
\ncline{->}{c}{b}
\ncline[offset=-3pt]{->}{c}{e}
\ncline[offset=3pt]{->}{c}{e}
\ncline[offset=-3pt]{->}{d}{c}
\ncline[offset=3pt]{->}{d}{c}
\ncline{->}{d}{f}
\ncline[offset=-3pt]{->}{e}{d}
\ncline[offset=3pt]{->}{e}{d}
\ncline{->}{e}{g}
\ncline{->}{E}{G}
\ncline{->}{F}{G}
}
\end{pspicture}
\end{minipage}
}
%
%
\subfigure[Inserting arrows from \Venus\ $-$ \texttt{A} to \texttt{D}. For each
inserted arrow the source is randomly selected from the distal boundary (shaded
nodes).] {
\begin{minipage}{\linewidth}
\centering
\begin{pspicture}[showgrid=false](7,3.5)
\psscalebox{.7}{
\rput(0,3){\circlenode{A}{$0$}}
\rput(1.5,4){\circlenode[fillstyle=solid,fillcolor=lightgray]{B}{$1$}}
\rput(1.5,2){\circlenode[fillstyle=solid,fillcolor=lightgray]{C}{$2$}}
\rput(3,3){\circlenode{b}{$1'$}}
\rput(4.5,3){\circlenode{c}{$2'$}}
\rput(6,4){\circlenode{d}{$3'$}}
\rput(6,2){\circlenode{e}{$4'$}}
\rput(6.75,5){\circlenode{f}{$5'$}}
\rput(6.75,1){\circlenode{g}{$6'$}}
\rput(8.25,4){\trinode[trimode=R,fillstyle=solid,fillcolor=lightgray]{E}{$4$}}
\rput(8.25,2){\circlenode[fillstyle=solid,fillcolor=lightgray]{F}{$5$}}
\rput(9.75,3){\circlenode[doubleline=true]{G}{$6$}}
\ncline{->}{A}{B}
\ncline{->}{A}{C}
\ncline{->}{c}{b}
\ncline[offset=-3pt]{->}{c}{e}
\ncline[offset=3pt]{->}{c}{e}
\ncline[offset=-3pt]{->}{d}{c}
\ncline[offset=3pt]{->}{d}{c}
\ncline{->}{d}{f}
\ncline[offset=-3pt]{->}{e}{d}
\ncline[offset=3pt]{->}{e}{d}
\ncline{->}{e}{g}
\ncline{->}{E}{G}
\ncline{->}{F}{G}
%
\ncline[linewidth=1.8pt]{->}{B}{b}
\nccurve[angleA=45,angleB=180,linewidth=1.8pt]{->}{B}{f}
\ncline[linewidth=1.8pt]{->}{F}{g}
}
\end{pspicture}
\end{minipage}
}
%
%
%
\subfigure[Inserting arrows from \texttt{D} to \Venus $-$ \texttt{A}. For each
inserted arrow the source is randomly selected from the proximal boundary
(shaded nodes).] {
\begin{minipage}{\linewidth}
\centering
\begin{pspicture}[showgrid=false](7,3.2)
\psscalebox{.7}{
\rput(0,3){\circlenode{A}{$0$}}
\rput(1.5,4){\circlenode{B}{$1$}}
\rput(1.5,2){\circlenode{C}{$2$}}
\rput(3,3){\circlenode[fillstyle=solid,fillcolor=lightgray]{b}{$1'$}}
\rput(4.5,3){\circlenode{c}{$2'$}}
\rput(6,4){\circlenode{d}{$3'$}}
\rput(6,2){\circlenode{e}{$4'$}}
\rput(6.75,5){\circlenode[fillstyle=solid,fillcolor=lightgray]{f}{$5'$}}
\rput(6.75,1){\circlenode[fillstyle=solid,fillcolor=lightgray]{g}{$6'$}}
\rput(8.25,4){\trinode[trimode=R]{E}{$4$}}
\rput(8.25,2){\circlenode{F}{$5$}}
\rput(9.75,3){\circlenode[doubleline=true]{G}{$6$}}
\ncline{->}{A}{B}
\ncline{->}{A}{C}
\ncline{->}{c}{b}
\ncline[offset=-3pt]{->}{c}{e}
\ncline[offset=3pt]{->}{c}{e}
\ncline[offset=-3pt]{->}{d}{c}
\ncline[offset=3pt]{->}{d}{c}
\ncline{->}{d}{f}
\ncline[offset=-3pt]{->}{e}{d}
\ncline[offset=3pt]{->}{e}{d}
\ncline{->}{e}{g}
\ncline{->}{E}{G}
\ncline{->}{F}{G}
%
\ncline{->}{B}{b}
\nccurve[angleA=45,angleB=180]{->}{B}{f}
\ncline{->}{F}{g}
\ncline[linewidth=1.8pt]{->}{f}{E}
\nccurve[angleA=165,angleB=-45,linewidth=1.8pt]{->}{g}{C}
\nccurve[angleA=0,angleB=-125,linewidth=1.8pt]{->}{g}{F}
\nccurve[angleA=55,angleB=-180,linewidth=1.8pt]{->}{g}{F}
}
\end{pspicture}
\end{minipage}
}
\caption{A concrete example of our crossover operator.
The numbers inside the nodes are labels for the nodes.}\label{fig:crossoverTwo}
\end{figure}

We give an outline of our crossover operator in Table~\ref{tab:crossover}. We
give a concrete example of our crossover operator in Figure~\ref{fig:crossoverTwo}.   As this example shows, the acceptor
and donor subgraphs can have different sizes; also, the two parents and the child can all have
different sizes.

The use of graph chromosomes is well established~\cite[p265]{banzhaf1998gpi}. Of
particular relevance to our work is research conducted by
Poli~\cite{poli1997evolution},~\cite{poli1996parallel}. He evolved computer programs
represented by graphs and he used the topology of his graphs to devise
evolutionary operators. Poli uses planar graphs, whereas our A-type graphs
need not be planar.  Poli's crossover operators exchanged connected subgraphs
of graphs of parents, as do our crossover operators, although we require that our
subgraphs be a radial set.
To our knowledge these graph-theoretic ideas have not previously been
used to devise evolutionary operators for A-types.

\subsection{Fitness Function}

 We use a standardized (and normalized) fitness function. That is, our fitness
 function returns a real number between 0 and 1, inclusive. The lower the fitness of an A-type, the more fit that A-type is. In this section we define our
fitness function.

Recall from the start of Section~\ref{sec:evolution} that we use our EA to
search for `suitably small' A-types that represent a particular function $f$.
We require \emph{training data} $T$ that is a set of input-output pairs of $f$.
That is, $T = \{ ( x_i , f(x_i) ) \}$ where $i$ is an element of some index
set. We call each pair in $T$ a \emph{training example}. We also require an
upper value $u$ for A-type sizes: A-types larger than $u$ are considered
unsuitable solutions.  We call $u$ the \emph{penalty bound}.  Note that in our algorithms, we always take the value of $u$ to be equal to the upper bound of the size of A-types in the initial population (see Section~\ref{sec:init_pop}).

Consider a candidate solution $A$. We determine the
fitness of $A$ as follows.
\begin{enumerate}
  \item \emph{Determining the performance of $A$ with respect to $T$.} Let
  $A(x_i)$ denote the output of $A$ given an input $x_i$. For each training
  example $( x_i , f(x_i) )$ we calculate the normalized Hamming distance between $A(x_i)$ and $f(x_i)$.
  Let $d$ denote the average of all of these Hamming distances.
  \item \emph{Including a penalty if $A$ is larger than $u$.} Choose a
  positive real number $m$, which we call the \emph{pressure gradient}. If
  $|A| \leq u$ then $A$'s fitness is $d$. Otherwise the fitness of $A$ is
  minimum of $\{ 1, d m ( |A| - u + 1 )\}$.
\end{enumerate}
Thus our fitness function is a continuous piecewise function $g$. It is
initially constant with $g = d$, then linear with a gradient $m$, then
constant with $g = 1$. This enables us to `pressure' the population so that
it is unlikely to contain A-types of size much greater than $u$.


\subsection{Selection Rules}

In our EA, for each generation we have three operations which require A-types
to be selected from the population: crossover, mutation, and elimination. In
this section we explain how we perform the selections.

For crossover our parent selection is a fitness proportional selection. The
fitter the A-type the greater the probability that that A-type is selected as a
parent. A-types are chosen by their fitness weighted by a function $h$; we chose
$h$ to be an exponential.  The choice of $h$ was the same for all the tasks we considered.

For the elimination operation A-types are also chosen by their fitness
weighted by an exponential. However, the less fit A-types are more likely to
be chosen for elimination.

For mutation our selection operator is random.

\subsection{Implementation}


\subsubsection{Candidate Solutions}

When our EA searches for an A-type $A$ that represents a given concept function
it has to search for both the graph of $A$ and the delay time $\delta$ of $A$.
For each candidate solution the EA chooses an A-type graph, estimates a range of possible
delays for that graph, and determines the fitness of each (graph + delay)
A-type.  That is, each candidate solution consists of a set of A-types all with the same
underlying A-type graph but with different delays coming from some interval.
So in our algorithm descriptions when we say that we make an A-type we are actually making
a set of A-types. We chose this implementation because it is easy to code and
efficient to run.

\subsubsection{Estimating a Range of Delays}

When our EA constructs an A-type graph $G$ (either
a randomly constructed graph for the initial population, or the result of
crossover or mutation) it must estimate a suitable range of delays for $G$.
Let $N$ denote the number of nodes in $G$. Let $A$ denote an A-type with graph
$G$ and a delay time $\delta = 0$. The larger the range of delays for each
individual, the longer it takes to train each individual. We take a somewhat
pragmatic approach to estimate the range of delays. To estimate the minimum
delay we perform the following four steps. First, we input a random sequence of vectors
into $A$. We
collect the output vectors from $A$ and call this sequence $S_{out}$. Second, we
repeat the above step, yielding a second output sequence $S_{out}'$. Third, we
determine the position $q$ where $S_{out}$ and $S_{out}'$ first differ (if
$S_{out} = S_{out}'$ then we set $q = -1$). This gives a reasonable estimate of the minimum possible time for data to percolate through the network from the input nodes to the output nodes.   Fourth, we subtract the sum of the
input dimension of $A$ and the output dimension of $A$ from $q$. If $q$ is
negative then we set it to zero. Our estimate of the minimum delay is $q$. We take the maximum delay to be $N$; this gives a reasonable estimate of the maximum possible time it can take data to percolate through the network from the input nodes to the output nodes.

\section{Simulations}\label{sec:simulations}

To investigate the performance of our EA we implemented the algorithm
using Java and ran many simulations with it. Here
we describe our simulations and present our results. Further detail can be found in~\cite{orr2010evolving}.

\subsection{Experimental Method}

Our simulations investigated the performance of our EA. We concentrated on two main
questions: whether our implementation of an evolutionary search is useful,
and whether our crossover operator aids our EA.

\subsubsection{Comparing Algorithms}

We compared three algorithms: a blind search, a
mutation-only EA, and an EA with crossover. First, we employed a blind search.
This algorithm simply creates a random A-type, and checks whether it is a
solution; if it is not then it is destroyed and the process is repeated. This is
a very special case of our EA; however, each candidate solution is entirely independent
of all previous candidate solutions---in the blind search all hereditary
information is lost from one generation to the next. Second, we employed a
mutation-only EA. Asexual evolution is seen in biology and it is a
straightforward special case of our EA---we simply ensure that no crossovers are
performed. Comparing our EA to the mutation-only special case offers a test of
the efficacy of our crossover operator. Third, we employed our EA in its entirety. We
name these algorithms \bsOne, \gaOneSans, and \gaOne\ respectively.

\subsubsection{Benchmark Learning Tasks}

To assess the performance of our EA we chose three simple supervised learning
tasks. These tasks involved searching for A-types
that represent simple classes of functions: $n$-identity, $n$-multiplexer and $n$-carry.
Their simplicity allowed us to investigate performance of our algorithm as the
complexity of the problem is scaled up. Also, it is easy to write down exact
solution A-types for each task investigated. In Sections~\ref{sec:searchingForIdentity},
\ref{sec:searchingForMux}, and \ref{sec:searchingForCarry} we describe
each task and the performance of our EA as it searches for that task. In this section
we give further details of our experimental method.

\subsubsection{What We Measured}

To gauge the performance of our algorithms we conducted several trials. For each
trial we recorded the
number of attempts required for a solution to be discovered: that
is, the total number of A-types constructed in the initial population,
via mutation and via crossover.  Note that \gaOneSans\ constructs one new A-type in each generation (by mutation), whereas \gaOne\ constructs two or more (by mutation and crossover).  For this reason we count the number of attempts rather than the number of generations.

Each learning task that we consider is a class of concept functions parametrised by a positive integer $n$ (usually $n$ is just the input dimension).  For each value of $n$ we
employed three algorithms and with each algorithm we conducted many trials. To
display our results we present a plot of $n$ versus attempts required.
A data point on these plots represents an average
of all trials for a particular algorithm searching for a particular concept
function. For all trials associated with one data point we employ Student's $t$-test (for instance see~\cite[sec
24.6]{kreyszig1994maths}) to determine a 90\% confidence interval. This
determines the error bars displayed around each data point. We assume that our
results are normally distributed, as is required for the $t$-test to be valid.

\subsubsection{Suitable Training Data}

Although we define A-types to accept and return sequential data, two of the
three concepts that we searched for are tasks that require A-types to be
operated in the clamped input mode. When we consider $n$-identity and
$n$-multiplexer concepts we do so with clamped examples. This makes our
investigations computationally easier. Conducting numerous trials with several
$n$ values is very computationally expensive if we search for A-types that
operate in the more general sequential mode. In order to test our EA with
A-types that operate in the sequential mode, we also devised a sequential task,
namely $n$-carry\footnote{Note that we use A-types with delay nodes for all
three learning tasks. However, it can be shown that there exist A-types without
delay nodes that represent clamped $n$-identity and clamped $n$-multiplexer; see Figure~\ref{fig:onetwoid} for $n=1$ and $n=2$.}.

Performing searches with long training
examples takes a long time; performing searches with short examples usually
leads to inexact solutions. Mindful of this we adopted the following procedure. We
chose relatively short training examples to discover possible solutions, then
tested these possible solutions with longer training examples. If a possible
solution fits these longer training examples then we deem it to be an
\emph{exact} solution (see below for more details).

When we searched for clampable A-types that represented a function $f$, we used a
training set containing all possible examples $(x,f(x))$ such that $f(x)$ is a
sequence of three vectors\footnote{With the exception of $n$-identity when
$n \in \{ 7,8,9,10 \}$; see
Section~\ref{sec:searchingForIdentity}.}. That is, when the fitness of a
candidate solution $A$ was assessed with an example $(x,f(x))$, the sequence
containing the first three Boolean vectors output by $A$ was compared with the
sequence $f(x)$. For example, Figure~\ref{fig:clampedTrainingSet} shows the
training set that we used when we searched for 2-identity.

\begin{figure}
\centering
\begin{minipage}{6cm}
\begin{tabular}{c}
\begin{minipage}{5cm}
\begin{displaymath}
 \left.%
	\left(%
	\phantom{a} \left(%
 	\twoVector{1}{1}
 	\right)\phantom{a},\phantom{a}
 	\left(%
 	\twoVector{1}{1},
 	\twoVector{1}{1},
 	\twoVector{1}{1}
 	\right) \phantom{a}
\right.%
\right)
\end{displaymath}
\end{minipage}
\\
\phantom{some vertical space}
\\
\begin{minipage}{5cm}
\begin{displaymath}
 \left.%
	\left(%
	\phantom{a} \left(%
 	\twoVector{1}{0}
 	\right)\phantom{a},\phantom{a}
 	\left(%
 	\twoVector{1}{0},
 	\twoVector{1}{0},
 	\twoVector{1}{0}
 	\right) \phantom{a}
\right.%
\right)
\end{displaymath}
\end{minipage}
\\
\phantom{some vertical space}
\\
\begin{minipage}{5cm}
\begin{displaymath}
 \left.%
	\left(%
	\phantom{a} \left(%
 	\twoVector{0}{1}
 	\right)\phantom{a},\phantom{a}
 	\left(%
 	\twoVector{0}{1},
 	\twoVector{0}{1},
 	\twoVector{0}{1}
 	\right) \phantom{a}
\right.%
\right)
\end{displaymath}
\end{minipage}
\\
\phantom{some vertical space}
\\
\begin{minipage}{5cm}
\begin{displaymath}
 \left.%
	\left(%
	\phantom{a} \left(%
 	\twoVector{0}{0}
 	\right)\phantom{a},\phantom{a}
 	\left(%
 	\twoVector{0}{0},
 	\twoVector{0}{0},
 	\twoVector{0}{0}
 	\right) \phantom{a}
\right.%
\right)
\end{displaymath}
\end{minipage}
\end{tabular}
\vspace{0.5cm}
\end{minipage}
\caption{The training set that we used for our searches for 2-identity. Note
that this training set is exhaustive in that this set contains all possible
examples of 2-identity that have output sequences with length $l = 3$.}
\label{fig:clampedTrainingSet}
\end{figure}

When we search for clampable A-types that represent a Boolean function $f$ we
define an exact solution as follows. Let $x_0$ denote a Boolean vector in the
domain of $f$. An A-type $A$ is an exact solution to $f$ if when
$x_0$ is input into $A$, the constant sequence $f(x_0)$ is returned by $A$ for $t$ moments, where $t$ is some
large but fixed positive integer. That is, the output nodes of $A$ have constant
value $f(x_0)$ for $t$ moments starting from moment $\delta$. When
searching for A-types that represent clamped $n$-identity and clamped
$n$-multiplexer, we deemed solutions to be exact when $t = 1000$.

When searching for A-types that represented a sequential function, our training data contained a single
example $(x,f(x))$, where $x$ was a random sequence of Boolean vectors. We chose $x$ to be short so that
solutions would often be found relatively quickly. As with the clamped case, to cater for
the chance that a discovered solution is incorrect we defined exact solutions for sequential
searches. We deemed a solution to be exact if it represents a training example
$(x,f(x))$ where $x$ consists of a random sequence of $10^4$ Boolean vectors.

\subsubsection{Other Search Parameters}

For each of the three algorithms tested there are several parameters that
require arguments; for instance, the population size, and
the mutation to crossover ratio. To optimize each algorithm we need to search
for appropriate arguments; furthermore, these arguments may be specific to each
benchmark concept. We performed some rather informal investigations to decide
upon arguments for these parameters. Those common to all three tasks are
presented in Table~\ref{table:commonArgs}. Further details are presented as we
introduce the investigations for each concept.
\begin{table}
\caption{Parameters common to all to three investigations$^\dagger$ in this section. Note
that the we set penalty bound $u$ equal to the upper bound for the size of A-types in the initial population.
This is specific to each learning task.}
\label{table:commonArgs}
\centering
\begin{tabular}{@{} p{6.5cm} l r @{}}
\midrule
\multicolumn{2}{c}{when using any algorithm}\\
parameter & argument \\
\midrule
population size & 100 \\
worst fitness of a solution  &  0.00 \\
probability that a node is constructed as a delay node & 20\% \\
penalty bound  & $u$ \\
pressure gradient & $\frac{1}{2}$ \\
\toprule
\multicolumn{2}{c}{when using \gaOne}\\
parameter & argument \\
\midrule
crossovers per generation  & 1 \\
mutations per generation  & 1 \\
upper bound of size (\% of internal nodes of parent) of donor or acceptor
subgraphs for crossover & 80\% \\
\bottomrule
\end{tabular}
\smallskip\\ $^\dagger\!$The number of crossovers and number of mutations per generation are allowed to vary in Section~\ref{sec:par_bias}.
\end{table}

\subsubsection{Task Management}\label{sec:taskManagement}

We conducted our investigations using many cores of numerous computers.
Consequently, we had to minimize any bias that this introduced into our
results. For each learning task we considered a set of concept functions, each of which
had a particular value of $n$. When we searched for a concept function $f$ we used
a set of a suitable number of training examples $\{ x , f(x) \}$. In the
cases where we did not use an exhaustive set of training examples, we randomly
selected a suitably sized training set from all possible examples. However, we
ensured that the training examples remained constant as the training algorithms
varied. That is, when we searched for an A-type representing a concept function $f$,
 the $i$th trial using each algorithm
had the same set of training examples. The processing time may vary from
computer to computer, but the number of attempts required should remain
constant.

For both $n$-identity and $n$-multiplexer, for each integer $n$ tested we
performed at least twenty trials for each algorithm. The exception to this
is for some blind searches, because on occasions the blind search took an excessively long
time to complete. We note below when twenty trials were not performed for the
blind search.

\subsection{Actual Solutions}

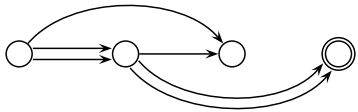
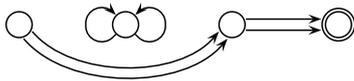
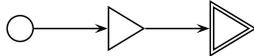
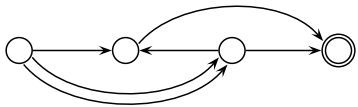
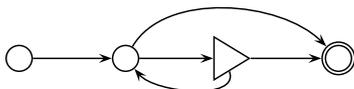
\begin{figure}
\begin{center}
\begin{tabular}{c}
\subfigure[The simplest A-type ($\delta = 2$) without delay nodes.]
{
\label{figure:actualIdentitySolns_a}
\begin{minipage}{0.4\textwidth}
\centering
\begin{pspicture}[showgrid=false](3,2)
\psscalebox{0.7}{
$
\rput(0,1){\circlenode{A}{\phantom{0}}}
\rput(2,1){\circlenode{B}{\phantom{0}}}
\rput(4,1){\circlenode[doubleline=true]{C}{\phantom{0}}}
\ncline[offset=3pt]{->}{A}{B}
\ncline[offset=-3pt]{->}{A}{B}
\ncline[offset=3pt]{->}{B}{C}
\ncline[offset=-3pt]{->}{B}{C}
$
}
\end{pspicture}
\\
\end{minipage}
}
\\
\subfigure[This A-type ($\delta = 2$) has a redundant node, namely the internal node without outgoing arrows. We may consider this node as `junk' because it does not contribute to the output of the A-type.]
{
\label{figure:actualIdentitySolns_b}
\begin{minipage}{0.4\textwidth}
\centering
\begin{pspicture}[showgrid=false](4.5,3)
\psscalebox{0.7}{
$
\rput(0,2){\circlenode{A}{\phantom{0}}}
\rput(2,2){\circlenode{B}{\phantom{0}}}
\rput(4,2){\circlenode{C}{\phantom{0}}}
\rput(6,2){\circlenode[doubleline=true]{D}{\phantom{0}}}
\ncline[offset=3pt]{->}{A}{B}
\ncline[offset=-3pt]{->}{A}{B}
\nccurve[angleA=50,angleB=130]{->}{A}{C}
\ncline{->}{B}{C}
\nccurve[offset=3pt,angleA=-50,angleB=-130]{->}{B}{D}
\nccurve[offset=-3pt,angleA=-50,angleB=-130]{->}{B}{D}
$
}
\end{pspicture}
\\
\end{minipage}
}
\\
\subfigure[This A-type ($\delta = 2$) also has a redundant node, namely the disconnected internal node.]
{
\label{figure:actualIdentitySolns_c}
\begin{minipage}{0.4\textwidth}
\centering
\begin{pspicture}[showgrid=false](4.5,2.5)
\psscalebox{0.7}{
$
\rput(0,2){\circlenode{A}{\phantom{0}}}
\rput(2,2){\circlenode{B}{\phantom{0}}}
\rput(4,2){\circlenode{C}{\phantom{0}}}
\rput(6,2){\circlenode[doubleline=true]{D}{\phantom{0}}}
\nccurve[offset=3pt,angleA=-50,angleB=-130]{->}{A}{C}
\nccurve[offset=-3pt,angleA=-50,angleB=-130]{->}{A}{C}
\ncdiag[angleA=-45,angleB=45,arm=0.8,linearc=.3]{->}{B}{B}
\ncdiag[angleA=-135,angleB=135,arm=0.8,linearc=.3]{->}{B}{B}
\ncline[offset=3pt]{->}{C}{D}
\ncline[offset=-3pt]{->}{C}{D}
$
}
\end{pspicture}
\\
\end{minipage}
}
\\
\subfigure[Although unnecessary, we include delay nodes in our searches for \emph{clamped} n-identity. Consequently we find solutions that involve delay nodes, as shown here ($\delta = 2$).]
{
\label{figure:actualIdentitySolns_d}
\begin{minipage}{0.4\textwidth}
\centering
\begin{pspicture}[showgrid=false](3,2)
\psscalebox{0.7}{
$
\rput(0,1){\circlenode{A}{\phantom{0}}}
\rput(2,1){\trinode[trimode=R]{B}{\phantom{0}}}
\rput(4,1){\trinode[doubleline=true,trimode=R]{C}{\phantom{0}}}
\ncline{->}{A}{B}
\ncline{->}{B}{C}
$
}
\end{pspicture}
\\
\end{minipage}
}
\\
\subfigure[The domain of a clamped function contains only constant sequences.
Consequently information may flow unsynchronised through an A-type solution;
this is the case for the A-type shown here ($\delta = 2$).] {
\label{figure:actualIdentitySolns_e}
\begin{minipage}{0.4\textwidth}
\centering
\begin{pspicture}[showgrid=false](4.5,3)
\psscalebox{0.7}{
$
\rput(0,2){\circlenode{A}{\phantom{0}}}
\rput(2,2){\circlenode{B}{\phantom{0}}}
\rput(4,2){\circlenode{C}{\phantom{0}}}
\rput(6,2){\circlenode[doubleline=true]{D}{\phantom{0}}}
\ncline{->}{A}{B}
\nccurve[offset=3pt,angleA=-50,angleB=-130]{->}{A}{C}
\nccurve[offset=-3pt,angleA=-50,angleB=-130]{->}{A}{C}
\nccurve[offset=-3pt,angleA=50,angleB=130]{->}{B}{D}
\ncline{->}{C}{B}
\ncline{->}{C}{D}
$
}
\end{pspicture}
\\
\end{minipage}
}
\\
\subfigure[An A-type ($\delta = 4$) whose graph has a closed directed path.]
{
\label{figure:actualIdentitySolns_f}
\begin{minipage}{0.4\textwidth}
\centering
\begin{pspicture}[showgrid=false](4.5,3)
\psscalebox{0.7}{
$
\rput(0,2){\circlenode{A}{\phantom{0}}}
\rput(2,2){\circlenode{B}{\phantom{0}}}
\rput(4,2){\trinode[trimode=R]{C}{\phantom{0}}}
\rput(6,2){\circlenode[doubleline=true]{D}{\phantom{0}}}
\ncline{->}{A}{B}
\ncline{->}{B}{C}
\nccurve[angleA=60,angleB=135]{->}{B}{D}
\nccurve[angleA=-70,angleB=-50]{->}{C}{B}
\ncline{->}{C}{D}
$
}
\end{pspicture}
\\
\end{minipage}
}
\end{tabular}
\caption{Some A-type solutions discovered when \gaOne\ was employed to
search for clamped identity.}  \label{figure:actualIdentitySolns}
\end{center}
\end{figure}

In this section we give examples of solutions obtained by our algorithms.  In Figure~\ref{figure:actualIdentitySolns} we present some of the solutions
found when we employed \gaOneSans\ to search for clamped identity function with one input and one output (clamped 1-identity function in the language of Section~\ref{sec:searchingForIdentity}).  The details of the search are given in Section~\ref{sec:searchingForIdentity}.  We found simple
solutions without delay nodes; see Figure~\ref{figure:actualIdentitySolns_a}. We
found solutions with subgraphs that did not contribute to the output of the
solutions; see Figures~\ref{figure:actualIdentitySolns_b}~and~\ref{figure:actualIdentitySolns_c}. Note that such subgraphs may be considered
`junk'; however, A-types with such subgraphs may prove to be useful intermediary
forms in an algorithm based on a population of A-types. In
Section~\ref{sec:clampedAtypes} we explained that there always
exists an A-type without delay nodes that represents a given clamped function.
However, for all simulations in this section we used A-types with delay nodes.
Consequently, we found solutions that involve delay nodes; see
Figure~\ref{figure:actualIdentitySolns_d}. Because A-types that represent
clamped functions do not require the synchronisation of data, we found
solutions having paths of differing lengths from the input node to the output node; see
Figures~\ref{figure:actualIdentitySolns_e}~and~\ref{figure:actualIdentitySolns_f}.

\subsection{Searching for Clamped $n$-Identity} \label{sec:searchingForIdentity}

\begin{figure}[!htp]
\label{fig:onetwoid}
\begin{center}
%
\subfigure[An A-type, with $\delta = 2$, that represents 1-identity.]
{
\label{fig:nIdentity_a}
\begin{pspicture}[showgrid=false](8,1.2)
\psscalebox{.7}{
\rput(3,1){\circlenode{A}{\phantom{0}}}
\rput(5,1){\circlenode{B}{\phantom{0}}}
\rput(7,1){\circlenode[doubleline=true]{C}{\phantom{0}}}
\ncline[offset=3pt]{->}{A}{B}
\ncline[offset=-3pt]{->}{A}{B}
\ncline[offset=3pt]{->}{B}{C}
\ncline[offset=-3pt]{->}{B}{C}
}
\end{pspicture}
}
%
%
\subfigure[An A-type, with $\delta = 2$, that represents 2-identity.]
{
\label{fig:nIdentity_b}
\begin{pspicture}[showgrid=false](8,2)
\psscalebox{.7}{
\rput(3,1){\circlenode{A}{\phantom{0}}}
\rput(3,2){\circlenode{B}{\phantom{0}}}
\rput(5,1){\circlenode{C}{\phantom{0}}}
\rput(5,2){\circlenode{D}{\phantom{0}}}
\rput(7,1){\circlenode[doubleline=true]{E}{\phantom{0}}}
\rput(7,2){\circlenode[doubleline=true]{F}{\phantom{0}}}
\ncline[offset=3pt]{->}{A}{C}
\ncline[offset=-3pt]{->}{A}{C}
\ncline[offset=3pt]{->}{B}{D}
\ncline[offset=-3pt]{->}{B}{D}
\ncline[offset=3pt]{->}{C}{E}
\ncline[offset=-3pt]{->}{C}{E}
\ncline[offset=3pt]{->}{D}{F}
\ncline[offset=-3pt]{->}{D}{F}
}
\end{pspicture}
}
\caption{Two A-types that represent the Boolean function $n$-identity for $n
\in \{1 , 2 \}$. Note that these also represent the more general function
columnwise $n$-identity.}
\label{fig:nIdentity}
\end{center}
\end{figure}
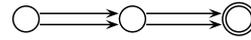
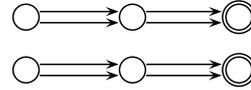

The first class of concept functions that we consider is clamped $n$-identity.
Given a positive integer $n$, $n$-identity is the Boolean function $f_{id}$ from $S_n$ to $S_n$
that maps each $n$-component Boolean vector to itself. In
Figure~\ref{fig:nIdentity} we illustrate two examples (found by inspection) of A-types that represent
 clamped $n$-identity---note that these also represent the more general function
 columnwise $n$-identity.

In this section we describe our searches for A-types that represent
$n$-identity for values of $n$ that range from 1 to 10. In
Table~\ref{table:argsForClampedIdentity} we list the arguments that we chose for
this search. When we searched for $n$-identity we employed all examples
with output sequences of length 3 unless there were more than 100 of these
(this was the case when $n \in \{7,8,9,10\}$). In the latter case we randomly
chose 100 examples for each trial.
Our choice of training data almost always gave exact solutions and, as described above,
we ensured that this choice was not a variable when we compared our algorithms.

\begin{table}
\caption{Parameters used for our clamped $n$-identity searches.}
\label{table:argsForClampedIdentity}
\centering
\begin{tabular}{@{}l r@{}}
\toprule
parameter & argument \\
\midrule
lower bound for size of initial machines  & $3n$ \\
upper bound for size of initial machines  & $4n$ \\
max num of attempts & $10^9$  \\
trials per training example  &  30 \\
length for exact solution & $10^3$ \\
\bottomrule
\end{tabular}
\end{table}

\begin{figure}
\includegraphics[width=\plotWidth]{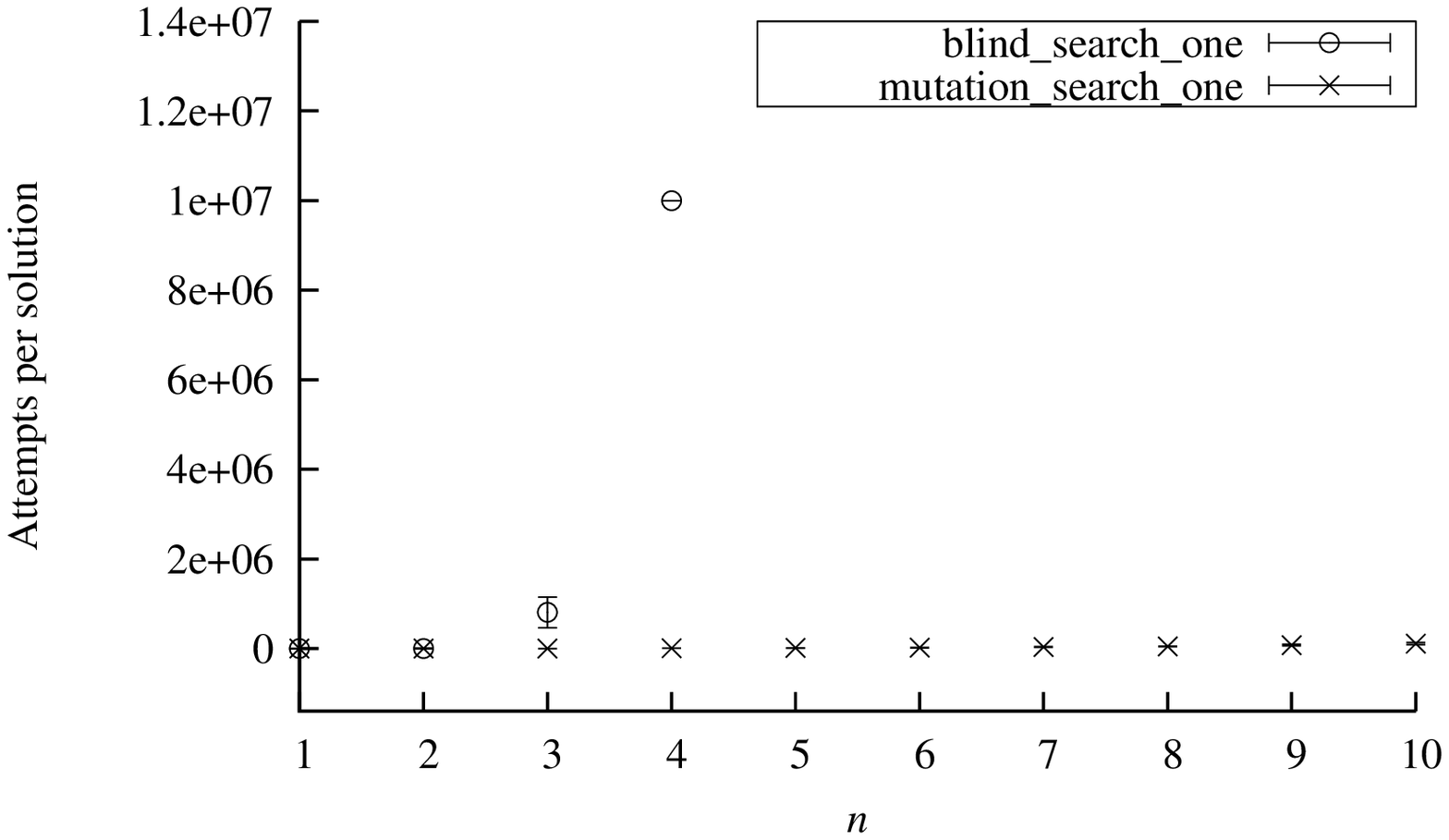}
\caption{Searching for A-types that represent $n$-identity with \bsOne\ and
\gaOneSans. Here we show the average number of attempts required before a solution was discovered.} \label{fig:mutVsBlind_identity_births}
\end{figure}

In Figure~\ref{fig:mutVsBlind_identity_births}
we compare the performances of \bsOne\ and
\gaOneSans\ when searching for A-types that represent $n$-identity with $n$ ranging from 1 to
10. Note that we do not display results for \bsOne\ for $n > 4$. This is because
all trials using \bsOne\ failed to find a solution within $10^9$ attempts.
These results show that \gaOneSans\ outperforms \bsOne\ by orders of magnitude.

\begin{figure}
\includegraphics[width=\plotWidth]{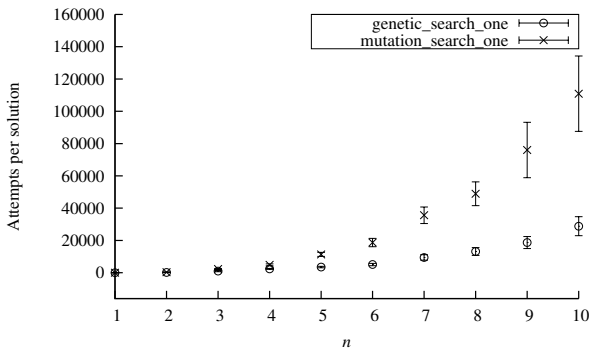}
\caption{Searching for A-types that represent $n$-identity with \gaOne\ and
\gaOneSans. Here we show the average number of attempts required before a solution was discovered.} \label{fig:mutVsCross_identity_births}
\end{figure}

In Figure~\ref{fig:mutVsCross_identity_births}
we compare the performances of
\gaOneSans\ and \gaOne\ when
searching for A-types that represent $n$-identity with $n$
ranging from 1 to 10. These results show that \gaOne\ significantly outperforms
\gaOneSans.

In conclusion, the results in this section provide evidence that when our EA
searches for $n$-identity it significantly outperforms a blind search. They also provide evidence
that when our EA searches for
clamped $n$-identity, our crossover operator aids our EA.

\subsection{Searching for Clamped $n$-Multiplexer}\label{sec:searchingForMux}

The second class of concept functions that we consider is clamped $n$-multiplexer. An
A-type $A$, with delay $\delta$, that represents $n$-multiplexer has $n$ regular input
nodes $(x_1, \ldots , x_n)$ and  $log_2(n)$ (rounded up to the next integer)
extra input nodes called {\em selector pins} $s_j$. Consider the input on the selector pins of $A$ at moment $t$.
This gives a binary representation of some integer $i$. At moment $(t + \delta)$ the
output of $A$ is equal to the value of $x_i$ at moment $t$.

Several researchers have applied
EAs to the task of discovering multiplexers. This started with
Wilson~\cite{wilson1987class} and others have also investigated this task, for
example Koza~\cite[ch 7]{koza1992genetic}, Butz~\cite[ch
3]{butz2002anticipatory}. In particular, Bull and
Preene~\cite{bull5481dynamical} used simulated evolution to design clampable
A-types that represent clamped $n$-multiplexers  Although
$n$-multiplexer is more complex than $n$-identity, it is another class of problem that scales easily.

\begin{figure}
\begin{center}
\subfigure[An A-type, with a delay $\delta = 3$, that represents 2-multiplexer.]
{
\label{fig:nMUX_a}
\begin{minipage}{\linewidth}
\centering
\begin{pspicture}[showgrid=false](5,5)
\psscalebox{.7}{
\rput(1,7){\circlenode{A}{\phantom{0}}}
\rput(1,5){\circlenode{B}{\phantom{0}}}
\rput(1,1){\circlenode{C}{\phantom{0}}}
\rput(2.5,7){\circlenode{D}{\phantom{0}}}
\rput(2.5,5){\trinode[trimode=R]{E}{\phantom{0}}}
\rput(2.5,3){\trinode[trimode=R]{F}{\phantom{0}}}
\rput(2.5,1){\trinode[trimode=R]{G}{\phantom{0}}}
\rput(4,6){\circlenode{H}{\phantom{0}}}
\rput(4,2){\circlenode{I}{\phantom{0}}}
\rput(5.5,4){\circlenode[doubleline=true]{J}{\phantom{0}}}
%
\rput(0.3,7){\circlenode[linecolor=white]{l1}{$s_0$}}
\rput(0.3,5){\circlenode[linecolor=white]{l2}{$x_0$}}
\rput(0.3,1){\circlenode[linecolor=white]{l3}{$x_1$}}
\rput(6.5,4){\circlenode[linecolor=white]{l4}{$y_0$}}
\ncline[offset=3pt]{->}{A}{D}
\ncline[offset=-3pt]{->}{A}{D}
\nccurve[angleA=-45,angleB=180]{->}{A}{F}
\ncline{->}{B}{E}
\ncline{->}{C}{G}
\ncline{->}{D}{H}
\ncline{->}{E}{H}
\ncline{->}{F}{I}
\ncline{->}{G}{I}
\ncline{->}{H}{J}
\ncline{->}{I}{J}
}
\end{pspicture}
\end{minipage}
}
\subfigure[An A-type, with a delay $\delta = 6$, that represents 3-multiplexer.]
{
\label{fig:nMUX_b}
\begin{minipage}{\linewidth}
\centering
\begin{pspicture}[showgrid=false](8,8)
\psscalebox{.7}{
\rput(1,9){\circlenode{A}{\phantom{0}}}
\rput(1,7){\circlenode{B}{\phantom{0}}}
\rput(1,3){\circlenode{C}{\phantom{0}}}
\rput(2.5,9){\circlenode{D}{\phantom{0}}}
\rput(2.5,7){\trinode[trimode=R]{E}{\phantom{0}}}
\rput(2.5,5){\trinode[trimode=R]{F}{\phantom{0}}}
\rput(2.5,3){\trinode[trimode=R]{G}{\phantom{0}}}
\rput(4,8){\circlenode{H}{\phantom{0}}}
\rput(4,4){\circlenode{I}{\phantom{0}}}
\rput(5.5,6){\circlenode{J}{\phantom{0}}}
\ncline[offset=3pt]{->}{A}{D}
\ncline[offset=-3pt]{->}{A}{D}
\nccurve[angleA=-45,angleB=180]{->}{A}{F}
\ncline{->}{B}{E}
\ncline{->}{C}{G}
\ncline{->}{D}{H}
\ncline{->}{E}{H}
\ncline{->}{F}{I}
\ncline{->}{G}{I}
\ncline{->}{H}{J}
\ncline{->}{I}{J}
\rput(1,11){\circlenode{a}{\phantom{0}}}
\rput(1,1){\circlenode{b}{\phantom{0}}}
\rput(2.5,11){\trinode[trimode=R]{c}{\phantom{0}}}
\rput(2.5,1){\trinode[trimode=R]{d}{\phantom{0}}}
\rput(4,10){\trinode[trimode=R]{e}{\phantom{0}}}
\rput(4,1){\trinode[trimode=R]{f}{\phantom{0}}}
\rput(5.5,10){\trinode[trimode=R]{g}{\phantom{0}}}
\rput(5.5,3){\trinode[trimode=R]{h}{\phantom{0}}}
\rput(7,10){\circlenode{i}{\phantom{0}}}
\rput(7,7){\trinode[trimode=R]{j}{\phantom{0}}}
\rput(7,5){\trinode[trimode=R]{k}{\phantom{0}}}
\rput(7,3){\trinode[trimode=R]{l}{\phantom{0}}}
\rput(8.5,8){\circlenode{m}{\phantom{0}}}
\rput(8.5,4){\circlenode{n}{\phantom{0}}}
\rput(10,6){\circlenode[doubleline=true]{o}{\phantom{0}}}
%
\rput(0.3,9){\circlenode[linecolor=white]{l1}{$s_0$}}
\rput(0.3,11){\circlenode[linecolor=white]{l2}{$s_1$}}
\rput(0.3,7){\circlenode[linecolor=white]{l3}{$x_0$}}
\rput(0.3,3){\circlenode[linecolor=white]{l4}{$x_1$}}
\rput(0.3,1){\circlenode[linecolor=white]{l5}{$x_2$}}
\rput(11,6){\circlenode[linecolor=white]{l6}{$y_0$}}
\ncline{->}{a}{c}
\ncline{->}{b}{d}
\nccurve[angleA=-30,angleB=150]{->}{c}{e}
\ncline{->}{d}{f}
\ncline{->}{e}{g}
\nccurve[angleA=30,angleB=-150]{->}{f}{h}
\ncline{->}{d}{f}
\ncline[offset=3pt]{->}{g}{i}
\ncline[offset=-3pt]{->}{g}{i}
\nccurve[angleA=-30,angleB=150]{->}{g}{k}
\ncline{->}{J}{j}
\ncline{->}{h}{l}
\ncline{->}{i}{m}
\ncline{->}{j}{m}
\ncline{->}{k}{n}
\ncline{->}{l}{n}
\ncline{->}{m}{o}
\ncline{->}{n}{o}
}
\end{pspicture}
\end{minipage}
}
\caption{A-types that represent $n$-multiplexer for $n \in \{ 2, 3 \}$.}
\label{fig:nMUX}
\end{center}
\end{figure}
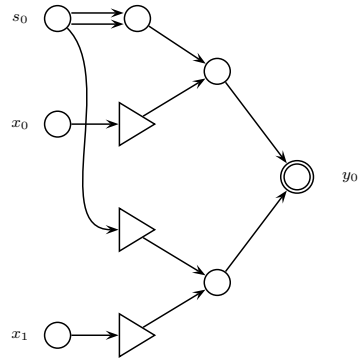
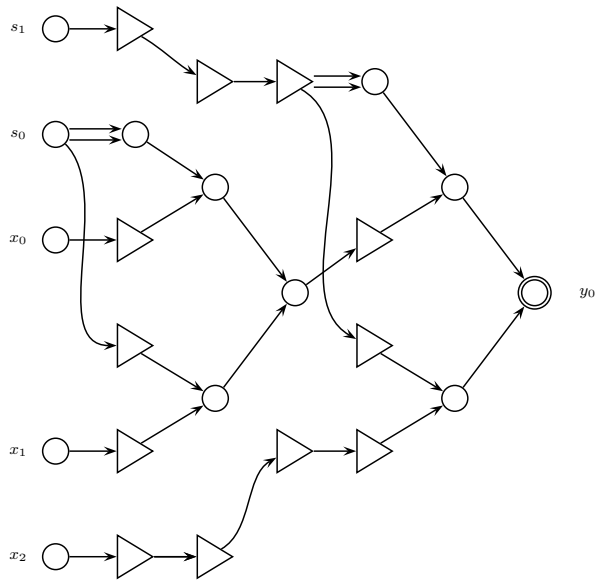

In this section we describe our searches for A-types that represent
$n$-multiplexer for values of $n$ that range from 2 to 5. In
Table~\ref{table:argsForClampedMux} we list the arguments that we chose for this
search. In Figure~\ref{fig:nMUX} we illustrate A-types (found by inspection) that represent
$n$-multiplexer where $n \in \{ 2 , 3 \}$---note that these also represent
columnwise $n$-multiplexer.

\begin{table}
\caption{Parameters used for our clamped $n$-multiplexer searches.
Note that for the lower bound we devised the following function $l$ of $n$.
$\{l(2) = 7$, $l(3) = 13$, $l(4) = 18$, $l(5) = 24\}$ by examining solutions
constructed by concatenating copies of our 2-multiplexer.}
\label{table:argsForClampedMux}
\centering
\begin{tabular}{@{}l r@{}}
\toprule
parameter & argument \\
\midrule
lower bound for size of initial machines  & $l(n)$ \\
upper bound for size of initial machines  & $l(n) + 4$ \\
max num of attempts & $10^8$  \\
trials per training example  &  20 \\
length for exact solution & $10^3$ \\
\bottomrule
\end{tabular}
\end{table}

\begin{figure}
\includegraphics[width=\plotWidth]{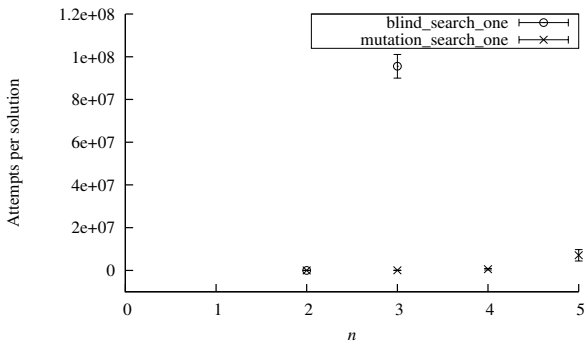}
\caption{Searching for A-types that represent $n$-multiplexer with \bsOne\ and
\gaOneSans. Here we show the average number of attempts required before a solution was discovered.}
\label{fig:mutVsBlind_mux_births}
\end{figure}

In Figure~\ref{fig:mutVsBlind_mux_births}
we compare the performances of \bsOne\ and \gaOneSans. When we used \bsOne\ to
search for A-types that represent $3$-multiplexer, only two of the twenty trials
returned a solution (before $10^8$ attempts). We include the data point
corresponding to $n = 3$ for \bsOne\ as a lower bound; that is, we expect that
had we allowed a greater maximum number of generations, the point
corresponding to 3-multiplexer for \bsOne\ would be greater than that shown.
These results show that \gaOneSans\ significantly out-performs \bsOne.

\begin{figure}
\includegraphics[width=\plotWidth]{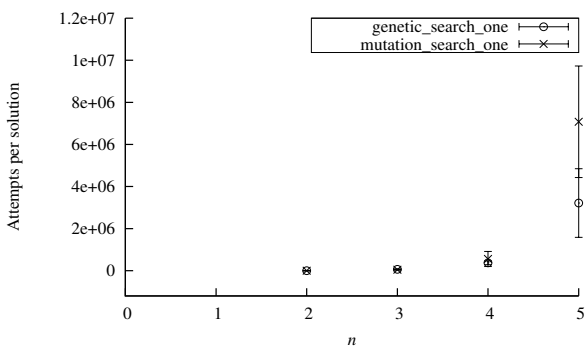}
\caption{Searching for A-types that represent $n$-multiplexer with \gaOne\ and
\gaOneSans. Here we show the average number of attempts required before
a solution was discovered.} \label{fig:mutVsCross_mux_births}
\end{figure}

In Figure~\ref{fig:mutVsCross_mux_births}
we compare the performances of \gaOneSans\ and \gaOne. These results show that when
searching for $n$-multiplexers for $n \in \{2,3,4,5\}$ there is no
conclusive difference between the performance of our two EAs. Considering the
relative positions of the means of the trials for 5-multiplexer, we speculate that as $n$ increases the crossover operator may prove to be
beneficial when our EA searches for $n$-multiplexers.

In conclusion, the results in this section provide evidence that when our EA
searches for $n$-multiplexer it significantly outperforms a blind search.
However, they fail to provide strong evidence that when
our EA searches for $n$-multiplexer our crossover operator aids our EA.

\subsection{Searching for Sequential $n$-Carry}\label{sec:searchingForCarry}

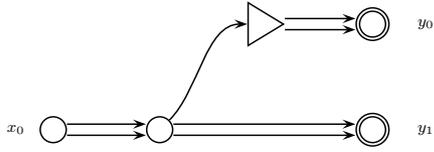
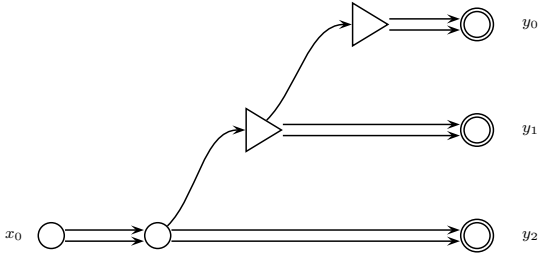
\begin{figure}
\begin{center}
\subfigure[An A-type, with a delay $\delta = 3$, that represents 2-carry.]
{
\centering
\begin{pspicture}[showgrid=false](6,3)
\psscalebox{.7}{
\rput(1,1){\circlenode{A}{\phantom{0}}}
\rput(3,1){\circlenode{B}{\phantom{0}}}
\rput(5,3){\trinode[trimode=R]{C}{\phantom{0}}}
\rput(7,3){\circlenode[doubleline=true]{D}{\phantom{0}}}
\rput(7,1){\circlenode[doubleline=true]{E}{\phantom{0}}}
%
\rput(0.3,1){\circlenode[linecolor=white]{l1}{$x_0$}}
\rput(8,3){\circlenode[linecolor=white]{l2}{$y_0$}}
\rput(8,1){\circlenode[linecolor=white]{l3}{$y_1$}}
\ncline[offset=3pt]{->}{A}{B}
\ncline[offset=-3pt]{->}{A}{B}
\nccurve[angleA=45,angleB=180]{->}{B}{C}
\ncline[offset=3pt]{->}{B}{E}
\ncline[offset=-3pt]{->}{B}{E}
\ncline[offset=3pt]{->}{C}{D}
\ncline[offset=-3pt]{->}{C}{D}
}
\end{pspicture}
}
\subfigure[An A-type, with a delay $\delta = 4$, that represents 3-carry.]
{
\centering
\begin{pspicture}[showgrid=false](8,4)
\psscalebox{.7}{
\rput(1,1){\circlenode{A}{\phantom{0}}}
\rput(3,1){\circlenode{B}{\phantom{0}}}
\rput(5,3){\trinode[trimode=R]{C}{\phantom{0}}}
\rput(7,5){\trinode[trimode=R]{D}{\phantom{0}}}
\rput(9,5){\circlenode[doubleline=true]{E}{\phantom{0}}}
\rput(9,3){\circlenode[doubleline=true]{F}{\phantom{0}}}
\rput(9,1){\circlenode[doubleline=true]{G}{\phantom{0}}}
%
\rput(0.3,1){\circlenode[linecolor=white]{l1}{$x_0$}}
\rput(10,5){\circlenode[linecolor=white]{l2}{$y_0$}}
\rput(10,3){\circlenode[linecolor=white]{l3}{$y_1$}}
\rput(10,1){\circlenode[linecolor=white]{l4}{$y_2$}}
\ncline[offset=3pt]{->}{A}{B}
\ncline[offset=-3pt]{->}{A}{B}
\nccurve[angleA=45,angleB=180]{->}{B}{C}
\ncline[offset=3pt]{->}{B}{G}
\ncline[offset=-3pt]{->}{B}{G}
\nccurve[angleA=45,angleB=180]{->}{C}{D}
\ncline[offset=3pt]{->}{C}{F}
\ncline[offset=-3pt]{->}{C}{F}
\ncline[offset=3pt]{->}{D}{E}
\ncline[offset=-3pt]{->}{D}{E}
}
\end{pspicture}
}
\caption{A-types that represent $n$-carry for $n \in \{2,3\}$}  \label{fig:nCarryOne}
\end{center}
\end{figure}

 In the previous two sections we searched for clampable A-types. The third class
 of concept functions that we consider is of a different kind to those
 previously considered: it does not consist of columnwise Boolean functions. We devised this class of functions to investigate a sequential
 task that has no clamped analogue.  We call this class of functions {\em $n$-carry}.
  Informally, $n$-carry maps a
 single bit string to a set of $n$ bit strings; each of these output strings is
 a segment of the input string. More formally, for some positive integer $n$
 and some integer $l \geq n$, $n$-carry is a function $f_n$
 from\footnote{Recall from Section~\ref{sec:repSeqFunctions} that we defined
 $S_{m,l}$ to denote the set of all sequences of length $l$ consisting of $m$-component
 Boolean vectors.} $S_{1,l}$ to  $S_{n,(l-n+1)}$, such that each sequence\\
\begin{displaymath}
x =
\left(%
[a_{l}], [a_{l-1}], \ldots ,  [a_1] %
\right)
\end{displaymath}
is mapped to\\
\begin{displaymath}
f_c(x) =
\left(%
\fourVector{a_{l-n+1}}{a_{l-n}}{\vdots}{a_l}, \ldots ,%
\fourVector{a_2}{a_3}{\vdots}{a_{n+1}},%
\fourVector{a_1}{a_2}{\vdots}{a_n}%
\right)
\end{displaymath}
For example in Figure~\ref{fig:nCarryExamples} we show four input-output pairs for 2-carry.  In
Figure~\ref{fig:nCarryOne} we illustrate two examples (found by inspection) of A-types that represent
 $n$-carry.
\begin{figure}
\centering
%
%
\begin{tabular}{c}
\begin{minipage}{8cm}
\begin{displaymath}
 \left.%
	\left(%
	\phantom{a} \left(%
 	\oneVector{1},\oneVector{0},\oneVector{1}
 	\right)\phantom{a},\phantom{a}
 	\left(%
 	\twoVector{0}{1},
 	\twoVector{1}{0}
 	\right) \phantom{a}
\right.%
\right),
\end{displaymath}
\end{minipage}
\\
\phantom{some vertical space}
\\
\begin{minipage}{8cm}
\begin{displaymath}
 \left.%
	\left(%
	\phantom{a} \left(%
 	\oneVector{1},\oneVector{1},\oneVector{1}
 	\right) \phantom{a} , \phantom{a}
  	\left(%
 	\twoVector{1}{1},
 	\twoVector{1}{1}
 	\right) \phantom{a}
\right.%
\right),
\end{displaymath}
\end{minipage}
\\
\phantom{some vertical space}
\\
\begin{minipage}{8cm}
\begin{displaymath}
 \left.%
	\left(%
	\phantom{a} \left(%
 	\oneVector{1},\oneVector{0},\oneVector{0}
 	\right) \phantom{a} , \phantom{a}
 	\left(%
 	\twoVector{0}{1},
 	\twoVector{0}{0}
 	\right) \phantom{a}
\right.%
\right),
\end{displaymath}
\end{minipage}
\\
\phantom{some vertical space}
\\
\begin{minipage}{8cm}
\begin{displaymath}
 \left.%
	\left(%
	\phantom{a} \left(%
 	\oneVector{0},\oneVector{0},\oneVector{1}
 	\right) \phantom{a} , \phantom{a}
 	\left(%
 	\twoVector{0}{0},
 	\twoVector{1}{0}
 	\right) \phantom{a}
\right.%
\right)
\end{displaymath}
\end{minipage}
\end{tabular}
\vspace{0.5cm}
%
\caption{Four input-output pairs of 2-carry.}  \label{fig:nCarryExamples}
\end{figure}

%

We searched for A-types that represent
$n$-carry for values of $n$ that range from 1 to 8. For each $n$-carry search
we chose a training example with a random input sequence of length 50. For each value of $n$
we conducted 20 trials per
algorithm and the training example for the $i$th trial was the same for all
algorithms. In Table~\ref{table:argsForCarry} we list the arguments that we chose
for this search.

\begin{table}
\caption{Parameters used for our sequential $n$-carry searches.}
\label{table:argsForCarry}
\centering
\begin{tabular}{@{}l r@{}}
\toprule
parameter & argument \\
\midrule
lower bound for size of initial machines & $3 + 2(n-1)$ \\
upper bound for size of initial machines  & $3 + 2n$ \\
max num of attempts & $10^9$  \\
trials per training example   &  $20^\dagger$ \\
length of exact solution & $10^4$ \\
\bottomrule
\end{tabular}
\smallskip\\$^\dagger10$ for the blind search.
\end{table}

\begin{figure}
\includegraphics[width=\plotWidth]{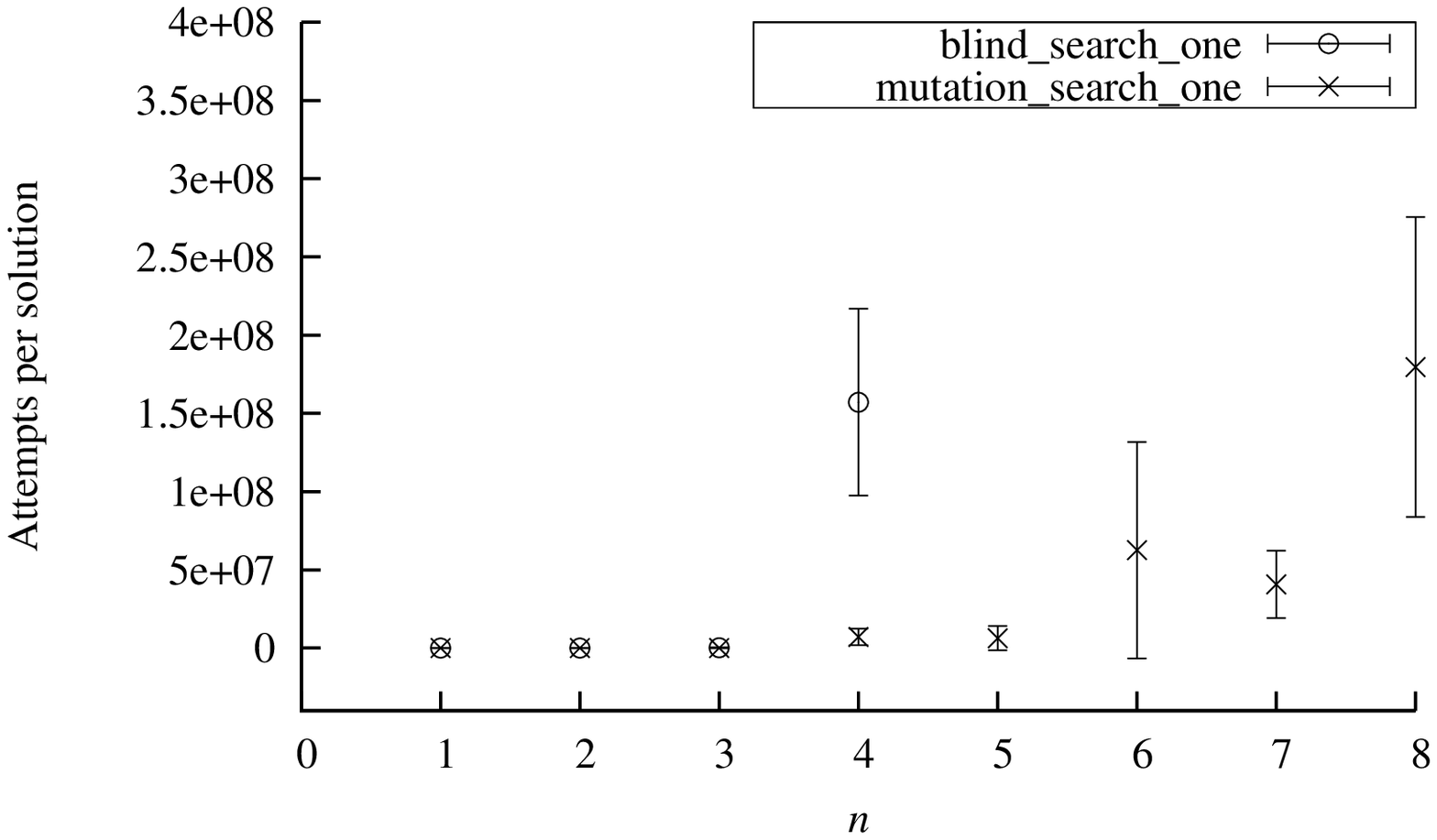}
\caption{Searching for A-types that represent $n$-carry with \bsOne\ and
\gaOneSans. Here we show the average number of attempts required before a solution was discovered.} \label{fig:mutVsBlind_carry_births}
\end{figure}

In Figure~\ref{fig:mutVsBlind_carry_births}
we compare the performances of \bsOne\ and \gaOneSans\ as they search for
$n$-carry, for $n$ ranging from 1 to 8. Note that when using \bsOne\, all trials for
$n > 4$ failed to find a solution.
From these two figures we
see that \gaOneSans\ significantly outperforms \bsOne.

\begin{figure}
\includegraphics[width=\plotWidth]{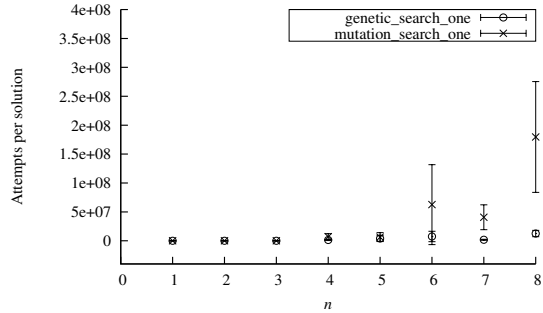}
\caption{Searching for A-types that represent $n$-carry with \gaOne\ and
\gaOneSans. Here we show the average number of attempts required before a solution was discovered.} \label{fig:mutVsCross_carry_births}
\end{figure}

In Figure~\ref{fig:mutVsCross_carry_births}
we compare the performances of \gaOne\ and \gaOneSans\ as they search for
$n$-carry. These results show that
\gaOne\ significantly outperforms
\gaOneSans.

In conclusion, the results in this section provide evidence that when our EA
searches for $n$-carry it significantly outperforms a blind search. They also
provide evidence that when our EA searches for $n$-carry our crossover operator
aids our EA.

\subsection{Parameter Bias}\label{sec:par_bias}

The above results suggest that our crossover operator is useful; however, we
must be mindful that \gaOne\ has many parameters that require arguments for a
particular search. Because our investigations were a `proof of concept' we
simply chose parameter values that ensured that we found solutions.
These values were held constant as we varied the algorithms.

\begin{figure}
\includegraphics[width=\plotWidth]{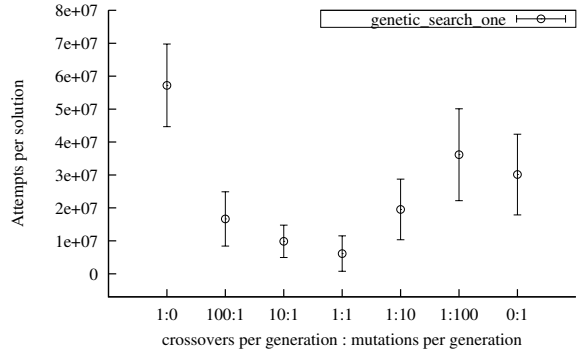}
\caption{Searching for A-types that represent $7$-carry using \gaOne\ and
various values of the ratio (crossovers per generation):(mutations per generation). Here we
show the average number of attempts required before a solution was discovered.}
\label{fig:sevenCarryRatioData}
\end{figure}

We did investigate the effect of varying the (crossovers per generation):(mutations per generation) ratio in \gaOne\ when searching for 7-carry.
The other parameter values for these simulations were those
specified in Tables~\ref{table:commonArgs}~and~\ref{table:argsForCarry}. The results are
presented in Figure~\ref{fig:sevenCarryRatioData}. Having a
(crossovers per generation):(mutations per generation) ratio of $1:1$ gave optimal performance.  Note that in the special case when the ratio is $0:1$, \gaOne\ is effectively the same as \gaOneSans.

\subsection{Is our Crossover Simply Macromutation?}

The above results provide evidence that our A-type crossover
operator is useful. However, we have yet to investigate whether this is simply because
our crossover operator is a `macromutator'; that is, whether our crossover
operator is only useful because it mixes the population
more effectively than our mutation operators. We turn to this question now.

The
results from the $n$-identity searches and the $n$-carry searches demonstrate
that for some tasks the crossover of our EA is useful. In many EAs crossover
is useful because it provides sudden large variation in the population, rather than
because it recombines individuals~\cite[ch 6]{banzhaf1998gpi}. Such an operator
is called a macromutation operator. This is not the case in biology: the utility
of \emph{biological} crossover is due to its ability to recombine individuals'
information~\cite[p276]{mackay2003information}.

The `headless chicken' search offers a relatively simple means of testing
whether a crossover operator is simply acting as a
macromutator~\cite{jones1995crossover}, ~\cite{poli2002exact}. The headless
chicken search is an EA where only one parent is selected from the population
and the other parent is an entirely new individual~\cite[p153]{banzhaf1998gpi}.
We implemented the headless chicken algorithm by duplicating \gaOne\ with the
following modification. For each crossover, after we have selected the parents
\Venus, \Mars\ we randomly choose one parent $P$ and then construct a random
A-type $P'$ that is the same size as $P$. We then perform the
crossover using $P'$ and the other parent.

\begin{figure}
\includegraphics[width=\plotWidth]{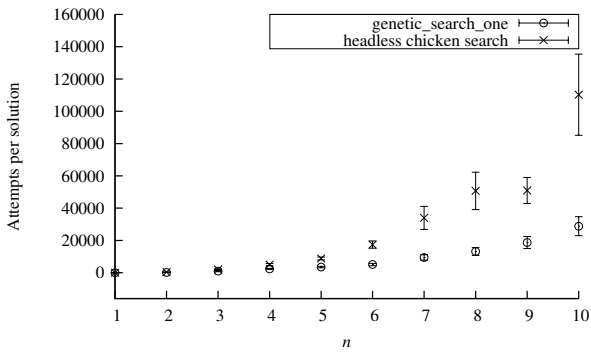}
\caption{Searching for A-types that represent $n$-identity with \gaOne\ and our headless
chicken crossover search. Here we show the average number of attempts required
before a solution was discovered.}
\label{fig:headlessVsTwoParents_identity_births}
\end{figure}

\begin{figure}
\includegraphics[width=\plotWidth]{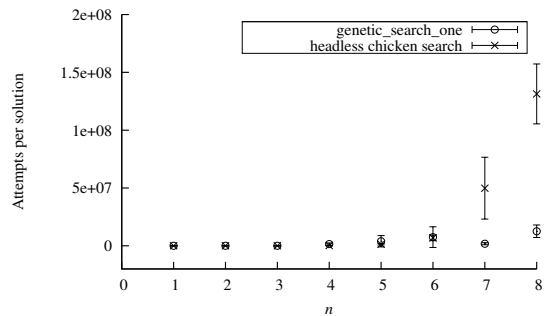}
\caption{Searching for A-types that represent $n$-carry with \gaOne\ and our
headless chicken crossover search. Here we show the average number of attempts
required before a solution was discovered.}
\label{fig:headlessVsTwoParents_carry_births}
\end{figure}

We compare \gaOne\ and our headless chicken search for clamped $n$-identity and $n$-carry,
the benchmark tasks that
demonstrated the utility of our crossover.
Figure~\ref{fig:headlessVsTwoParents_identity_births}
shows that when searching for
clamped $n$-identity, \gaOne\ outperforms our headless chicken search.
Figure~\ref{fig:headlessVsTwoParents_carry_births}
shows that when searching for
$n$-carry, \gaOne\ also outperforms our headless chicken search.

This provides evidence that for some tasks our crossover operator is more useful
than a macromutation operator.

\subsection{Size Bias}

\begin{figure}
\includegraphics[width=\plotWidth]{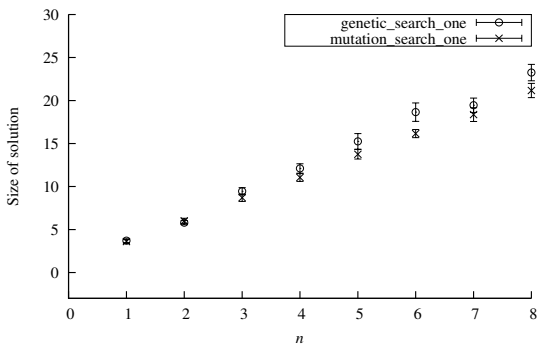}
\caption{The sizes of A-types found to represent $n$-carry using \gaOneSans\ and
\gaOne. Again, the error bars show the $90$\% confidence interval using
Student's t-test.}
\label{fig:solutionSizes}
\end{figure}

We now briefly turn to the size of solutions obtained by different algorithms.  Consider $n$-carry, for example.  The graph in
Figure~\ref{fig:solutionSizes} shows that there is not a great difference between the solution
sizes found by \gaOneSans\ and \gaOne.  Hence the difference in performance of these algorithms is not due to size differences in the populations.

More generally, one can consider the diversity of the population as the algorithm progresses.  It can be seen from Figure~\ref{fig:solutionSizes} that the algorithms found solutions of different sizes for each fixed value of $n$; in particular, these solutions were not all the same.  This indicates the presence of at least some diversity.  We did not investigate
population diversity systematically.  See also
Figure~\ref{figure:actualIdentitySolns}, which shows a sample of solutions
obtained by using \gaOneSans\ to search for A-types that represent $1$-identity.

Recall from Section~\ref{sec:evoOps} that our method for fitness-based
selection employs an exponential function. This strongly favours fitter individuals,
which may reduce the diversity of the population.  Our choice of exponential sufficed for our algorithm comparisons.  One advantage of our method for fitness-based selection is that it would be easy to
vary: one can replace the exponential with any other monotone function.

\section{Conclusion}

We devised a graph-based EA for finding A-types that represent a given function.
When applied to the three benchmark problems, the EA performed considerably better than
a purely random search.  For clamped $n$-identity and $n$-carry, the full version of the
EA performed better than the mutation-only version.  Our algorithm worked in both the
clamped and the sequential settings.

We now suggest directions for future research. A-types are
relatively simple, yet they are recurrent Boolean ANNs capable of representing
any Boolean function and operating in a sequential mode. Consequently, we believe A-types
are a useful tool for investigating the learning and behaviour of Boolean ANNs. In
particular, the simplicity of A-types means that manipulations of their graphs
are often straightforward to implement. We suggest two areas of future research with A-types: further
investigations into evolutionary techniques, and using the symmetry
of a concept function to improve the search for an A-type that represents that
function.

\subsection{Evolving Evolutionary Operators}\label{sec:evolving_ops}

Here we propose that it is worthwhile to continue to search for useful
evolutionary operators for A-types. Furthermore, we propose that evolutionary
searches can be applied to discover these operators. The evolution of
parameters of a search is an established technique in evolutionary
computing~\cite[ch 4]{eiben2003iec}. Many researchers have extended this idea
to include the evolution of evolutionary operators~\cite{tang2009biologically}.
 In terms of evolving networks, Teller's
research~\cite{teller1995program}~\cite{teller1996evolving} is of particular
interest. Teller solved signal classification tasks by evolving two populations
simultaneously. One population was a set of programs, which were
represented with graphs. The other population was a set of evolutionary
operators that operated on the programs. We believe that it would be worthwhile
to co-evolve evolutionary operators in a manner analogous to Teller's research.  This would allow a more complete investigation of what happens when one varies the many parameters in our EA.

The results in Section~\ref{sec:simulations} show that, for some
problems, our crossover operator is more useful than a macromutation operator. Although our
crossover operator employs relatively simple graph-theoretical ideas, its
implementation is rather involved.
By evolving evolutionary operators for A-types, one may be able to find more complicated but better-performing A-type crossover operators
and test whether certain properties (such as the out-degree of nodes, connectedness of subgraphs,
network activity\footnote{Loosely, we can define the activity of a node as the
average number of changes of state per moment it undergoes when a large random data packet is
processed by the network. Furthermore, we can define the activity of a
subgraph of an A-type as an average of the activity of all nodes in that
network. Note that Teuscher~\cite[ch 5]{teuscherOne} defines activity of
A-types and uses this to investigate the non-linear dynamics of these
networks.}, and perhaps some measure of symmetry) are useful.

\subsection{Making Use of Symmetry}

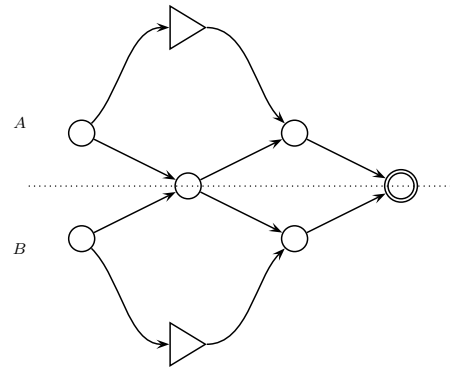
\begin{figure}
\begin{center}
\label{fig:mirrorSymmXOR_b}
\begin{pspicture}(4,5)
\psscalebox{.7}{
$
\rput(1,4){\circlenode{A}{\phantom{0}}}
\rput(1,2){\circlenode{B}{\phantom{0}}}
\rput(3,6){\trinode[trimode=R]{C}{\phantom{0}}}
\rput(3,0){\trinode[trimode=R]{D}{\phantom{0}}}
\rput(3,3){\circlenode{E}{\phantom{0}}}
\rput(5,4){\circlenode{F}{\phantom{0}}}
\rput(5,2){\circlenode{G}{\phantom{0}}}
\rput(7,3){\circlenode[doubleline=true]{H}{\phantom{0}}}
\psline[linestyle=dotted,dotsep=2pt](0,3)(8,3)
\nccurve[angleA=45,angleB=-180]{->}{A}{C}
\ncline{->}{A}{E}
\nbput[npos=-0.8]{A}
\ncline{->}{B}{E}
\naput[npos=-0.8]{B}
\nccurve[angleA=-45,angleB=-180]{->}{B}{D}
\nccurve[angleA=0,angleB=135]{->}{C}{F}
\ncline{->}{E}{F}
\ncline{->}{E}{G}
\nccurve[angleA=0,angleB=-135]{->}{D}{G}
\ncline{->}{F}{H}
\ncline{->}{G}{H}
$
}
\end{pspicture}
\end{center}
\caption{Redrawing the A-type, $\delta = 3$, shown in Figure~\ref{fig:XOR} to
emphasise the mirror symmetry of the solution.} \label{fig:mirrorSymmXOR}
\end{figure}

The notion of symmetry, which is made precise by group theory, leads to useful
problem-solving techniques. Consider the A-type shown in
Figure~\ref{fig:XOR}, which represents columnwise \xor. This function is
symmetric in its arguments: that is, $A \oplus B = B \oplus A$ for all $A$ and $B$. In
Figure~\ref{fig:mirrorSymmXOR} we redraw this A-type to show that it
has `mirror symmetry' about a horizontal line. So columnwise \xor\ has a symmetry;
when searching for an A-type that represents it, both the concept function and one of its solutions share this property. We hypothesise that this idea can be
formalised using group theory for a class of concept functions admitting a symmetry and used to cut down the size of the search space of an EA.  This is work in progress.

Recently Kondor~\cite{kondor2008group} investigated the use of group-theoretic methods to
improve some modern machine learning techniques. Other researchers have also applied
symmetries to ANNs for this
purpose~\cite{baldi1987symmetries}~\cite{shawe1993symmetries}~\cite{wood1996unifying}.
Recently Dong and Zhang~\cite{dong2006detection} incorporated group-theoretic
techniques into EAs with populations of ANNs, using relatively simple operators. The simplicity of
A-types makes them a good setting in which to further implement and test the application of group-theoretic ideas on a population of evolving ANNs.

\begin{acknowledgements}
The first author wishes to thank the Department of Physics and Astronomy, University of
Canterbury, for financial support.  We thank the referees for their thorough reading of the paper and their helpful suggestions for improving it.  We are grateful to Jack Copeland for many stimulating
discussions, Paul Brouwers and Orlon Petterson for technical support, Larry Bull for
helpful comments, and Mike Reid for general advice and encouragement.  Nikolai Kruetzmann
gave valuable feedback on an earlier draft of this paper.
\end{acknowledgements}

\bibliographystyle{spmpsci}      
\bibliography{papersBib} 

\end{document}